%% file: main.tex
\documentclass[10pt]{article} % For LaTeX2e
\usepackage[preprint]{tmlr}

\input{math_commands.tex}

\usepackage{hyperref}
\usepackage{url}

\usepackage{amsmath,amsfonts}
\usepackage{algcompatible}
\usepackage{algorithm}
\usepackage{array}
\usepackage{textcomp}
\usepackage{stfloats}
\usepackage{url}
\usepackage{verbatim}
\usepackage{graphicx}
\usepackage{cite}
\usepackage{booktabs} 
\usepackage{multirow}
\usepackage{xcolor}
\usepackage[super]{nth}
\usepackage{tikz}
\newcommand*\circled[1]{\tikz[baseline=(char.base)]{
            \node[shape=circle,draw,inner sep=0.2pt] (char) {#1};}}
\newcommand*\circledB[1]{\tikz[baseline=(char.base)]{
            \node[shape=circle,fill,inner sep=0.2pt] (char) {\textcolor{white}{#1}};}}
\usepackage{soul}
\usepackage{enumitem}
\usepackage{array}
\newcolumntype{?}{!{\vrule width 1.5pt}}
\usepackage{caption}
\usepackage{hyperref}

\title{Enabling Energy-Efficient Simultaneous Multi-Task Reinforcement Learning through Spiking Neural Networks with Active Dendrites for Bio-inspired Generalist Agents}

\author{\name Rachmad Vidya Wicaksana Putra\thanks{Equal Contributions} \email rachmad.putra@nyu.edu \\
      \addr eBRAIN Lab\\
      New York University (NYU) Abu Dhabi
      \AND
      \name Avaneesh Devkota\footnotemark[1] \email ad5768@nyu.edu  \\
      \addr eBRAIN Lab\\
      New York University (NYU) Abu Dhabi
      \AND
      \name Muhammad Shafique \email muhammad.shafique@nyu.edu\\
      \addr eBRAIN Lab\\
      New York University (NYU) Abu Dhabi}

\def\month{MM}  % Insert correct month for camera-ready version
\def\year{YYYY} % Insert correct year for camera-ready version
\def\openreview{\url{https://openreview.net/forum?id=XXXX}} % Insert correct link to OpenReview for camera-ready version

\begin{document}

\maketitle

%%%%%%%%%%%%%%%%%%%%%%%%%%%%%%%%%%%%%%%%%%%%%%%%%%%%%%%%%%%%%%%%%%%%%%%%%%%%%%%%%%%%%%%%%%%%%
%%%%%%%%%%%%%%%%%%%%%%%%%%%%%%%%%%%%%%%%%%%%%%%%%%%%%%%%%%%%%%%%%%%%%%%%%%%%%%%%%%%%%%%%%%%%%

\begin{abstract}
Reinforcement learning (RL) has demonstrated remarkable capabilities in training agents to solve complex tasks autonomously, such as mobile robots, UAVs/UGVs, and game-playing agents). 
However, scaling RL to master multiple tasks simultaneously (i.e., so-called \textit{multi-task RL}) remains a significant challenge.
Such a multi-task RL capability especially is important for agents to adapt to changes in real-world operational environments.
State-of-the-art works show that, training agents with neural networks and shared structures across tasks promises improved generalization in simultaneous multi-task RL. 
However, they still suffer from \textit{task interference} and incur high energy consumption due to intensive computation. 
To address this, we propose \textit{\textbf{MTSpark}}, a novel methodology that enables energy-efficient simultaneous multi-task RL using spiking neural networks (SNNs) equipped with active dendrites for bio-inspired generalist agents.
Specifically, MTSpark enhances a Deep Spiking Q-Network (DSQN) with active dendrites, a dueling structure, and task-specific context signals to dynamically form specialized sub-networks for individual tasks, while exploiting sparse operations for energy-efficient network processing. 
Experimental results demonstrate that MTSpark achieves higher performance and efficiency compared to state-of-the-art by obtaining high scores across three Atari games (i.e., Pong: -5.4, Breakout: 0.6, and Enduro: 371.2), approaching human-level performance (i.e., Pong: -3, Breakout: 31, Enduro: 368), while incurring similar memory and about 2x lower energy than state-of-the-art. 
These results show that our MTSpark potentially advances the frontiers toward energy-efficient generalist agents by combining RL and SNNs. 
\end{abstract}

%%%%%%%%%%%%%%%%%%%%%%%%%%%%%%%%%%%%%%%%%%%%%%%%%%%%%%%%%%%%%%%%%%%%%%%%%%%%%
%%%%%%%%%%%%%%%%%%%%%%%%%%%%%%%%%%%%%%%%%%%%%%%%%%%%%%%%%%%%%%%%%%%%%%%%%%%%%
\section{Introduction}
\label{sec_intro}

In recent years, reinforcement learning (RL) has emerged as a powerful approach for enabling agents to autonomously acquire complex behaviors. While RL methods have demonstrated exceptional success in single-task environments, such as mastering Chess, Go, and Atari games~\citep{Playing_Atari, Human_Level_Control, Mastering_Chess_Shogi}, scaling these techniques to simultaneous multi-task settings remains an open challenge~\citep{Zhang2018AnOOA}. 
The ability to train an agent on multiple tasks simultaneously is essential for dynamic, real-world applications that require robust generalization and adaptation~\citep{mt_rl}.
One of the key challenges is to manage and effectively use the increased amount of data from different tasks and extended training time~\citep{Ref_Espeholt_IMPALA_ICML18, Ref_Hessel_PopArt_AAAI19}.

\begin{figure}[t]
    \centering
    \includegraphics[width=\linewidth]{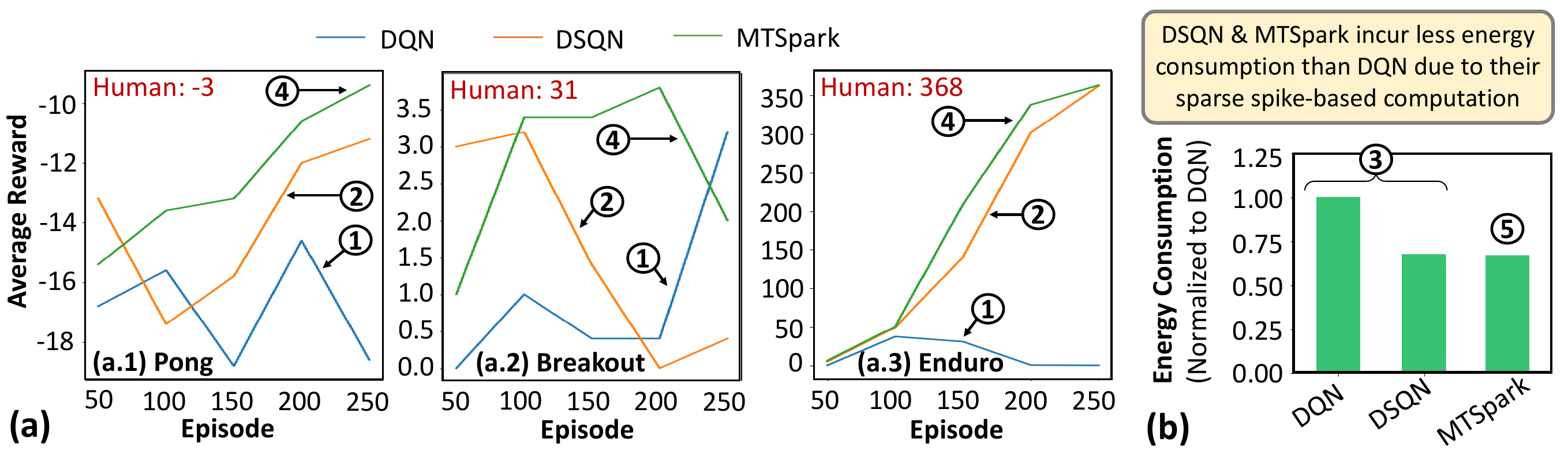}
    \vspace{-0.5cm}
    \caption{\textbf{(a)} Multi-task learning performance of the state-of-the-art RL-based methods for DNN (i.e., DQN) and SNN (i.e., DSQN), and our MTSpark, across three Atari games (i.e., Pong, Breakout, and Enduro). Our MTSpark achieves high performance in all tasks, while the state-of-the-art only perform well in certain tasks.
    \textbf{(b)} Estimated energy consumption of different RL-based methods: DQN, DSQN, and our MTSpark.}
    \label{fig_Observation}   
\end{figure}

A natural approach to multi-task RL is to train agents jointly on all tasks, leveraging shared structure and paramaters across tasks to improve sample efficiency and performance, as compared to solving each task independently~\citep{Sodhani2021MultiTaskRLA, DEramo2020SharingKIA}. 
However, this approach still suffers from \textit{task interference} challenges, where conflicting task objectives disrupt the learning process, resulting in suboptimal representations and leading to degraded performance~\citep{parisotto2016actormimicdeepmultitasktransfer, rusu2016policydistillation}.
These limitations are shown by labels-\circled{1} and \circled{2} in Figure~\ref{fig_Observation}(a), where the foundation RL-based methods in both Deep Neural Network (DNN) and Spiking Neural Network (SNN) domains struggle to effectively learn multiple tasks (i.e., three Atari games: Pong, Breakout, and Enduro). 
Therefore, Deep Q-Network (DQN)~\citep{Playing_Atari} and Deep Spiking Q-Network (DSQN)~\citep{DSQN} can only perform well in certain games. 
Furthermore, this approach also suffers from \textit{high energy consumption} due to its intensive computation, as shown in Figure~\ref{fig_Observation}(b).
These limitations severely impact the practicality of RL-based agents in real-world applications, where energy-efficient solutions with capabilities of adaptability and generalization across tasks are critically required (e.g., autonomous robots that can adapt to diverse environments)~\citep{mt_rl, Ref_SpikeDyn, Ref_lpSpikeCon, Ref_Minhas_NCLsurvey_Access25}.
Therefore, the \textbf{targeted research problem} is \textit{how can we develop energy-efficient RL process that can learn multiple tasks simultaneously without significant performance degradation for generalist agents?}
An efficient solution to this problem is a step forward in realizing general artificial intelligence (AI).

%%%%%%%%%%%%%%%%%%%%%%%%%%%%%%%%%%%%%%%%%%%%%%%%%%%
\subsection{State-of-the-Art and Their Limitations}
\label{sec_intro_sota}

\textbf{DNN Domain:}
Several strategies have been proposed, each striving to balance the shared knowledge with the task-specific information.
\textit{Replay-based methods} aim to strengthen the knowledge of previously learned tasks by replaying the corresponding data~\citep{Replay2, Ref_Minhas_NCLsurvey_Access25, Ref_Minhas_Replay4NCL_DAC25, Replay1}.
However, they come with significant memory and storage demands to keep large amounts of previous data, thus limiting their scalability. 
\textit{Architectural-based methods} employ modular architectures to separate task-specific components from shared representations. 
However, they also face scalability issues, since adding new components lead to uncontrolled growth in network size and compute cost~\citep{Ref_Minhas_NCLsurvey_Access25}. 
Prominent works in RL-based DNNs introduce DQN architecture~\citep{Playing_Atari}, which performs exceptionally well in single-task settings but struggles in multi-task settings; see \circled{1} of Figure~\ref{fig_Observation}(a).
Specifically, DQN achieves notable performance improvement in Breakout after 200 training episodes\footnote{An episode is a full game playthrough where the agent interacts with the environment until the game ends.}, while struggling to master other games.  
This shows the difficulty of applying RL models to simultaneous multi-task settings.
Recently, IMPALA performs multi-task learning by employing \textit{actors} to generate actions in environments, and \textit{a learner} to train a central model on the collected data~\citep{Ref_Espeholt_IMPALA_ICML18}.
Meanwhile, PopArt complements IMPALA by stabilizing the reward scale changes across tasks~\citep{Ref_Hessel_PopArt_AAAI19}.
These works~\citep{Ref_Espeholt_IMPALA_ICML18, Ref_Hessel_PopArt_AAAI19} have demonstrated improvements in multi-task RL but they still rely on intensive processing, which incurs high energy consumption, thereby making it unsuitable for resource-constrained autonomous agents.

\textbf{SNN Domain:} 
Bio-inspired SNNs have gained traction as promising alternative energy-efficient machine learning (ML) algorithms, since they show substantial potential in diverse applications~\citep{Active_Image, Ref_Putra_FSpiNN_TCAD20, Ref_Cordone_ObjDetSNN_IJCNN22, Ref_Luo_EEGSNN_Access20, Ref_Bartolozzi_EmbodiedNeuroIntel_Nature22, Ref_Putra_SpikeNAS_TAI26}. 
The inherent ability of SNNs to process temporal information efficiently makes them particularly suitable for data stream processing, a key feature for enabling multi-task learning. 
For instance, TM-SNN employs threshold modulated SNNs for learning multiple tasks, but these tasks are based on a small NMNIST dataset~\citep{Ref_Cachi_TMSNN_IWANN23}. 
Hence, its performance has not been evaluated for more complex tasks, such as Atari games.
Recently, DSQN~\citep{DSQN}, the SNN equivalent of DQN, has higher performance than DQN in both Pong and Enduro after 150 training episodes, while incurring lower energy consumption than DQN, as shown by labels-\circled{2} and \circled{3} in Figure~\ref{fig_Observation}.  
These results highlight the potentials of employing SNNs to exploit temporal information and sparse computation for improving performance and efficiency of multi-task RL as well as enabling on-device learning.
However, \textit{the potential of SNNs in on-device multi-task RL remains largely unexplored, hindering their applicability in real-world use-cases where generalization is crucial}.
Moreover, many multi-task RL methods are not suitable as solutions, as they rely on separate task training and testing. 
Meanwhile, \textit{the targeted problem is simultaneous multi-task RL where different tasks are learned at the same time, directly addressing multiple tasks with only one network model}.

% %%%%%%%%%%%%%%%%%%%%%%%%%%%%%%%%%%%%%%%%%%
\subsection{Associated Research Challenges}
\label{sec_intro_challenges}

Developing an RL-based SNN architecture that can support multi-task learning is non-trivial.  
The associated \textbf{key research challenges} include the following.
\begin{itemize}[leftmargin=*]
    \item The network should employ an effective yet simple function to quantify the reward and punishment values for training process, enabling an efficient RL mechanism.
    \item The network should leverage task-specific context information (e.g., task identity) to effectively guide the learning process.
    \item The network needs to understand the importance of different actions and their consequences after each state transition, since the agent should identify the most appropriate action to take based on the state and the task context, optimizing performance for each individual task while maintaining efficiency across different task
\end{itemize}

% %%%%%%%%%%%%%%%%%%%%%%%%%%%%%%%%%%%%%%%%%%
\subsection{Our Novel Contributions}
\label{sec_intro_novelty}

To address the research challenges, we propose \textbf{MTSpark}, \textit{a novel methodology to enable energy-efficient simultaneous \underline{M}ulti-\underline{T}ask reinforcement learning using \underline{Sp}iking neur\underline{a}l netwo\underline{rk}s for realizing bio-inspired generalist agents}; see the overview in Figure~\ref{fig_methodology}.
Moreover, it is also the first work that enables simultaneous multi-task RL through SNNs, marking a notable step toward efficient, bio-inspired generalist agents.
Its key novel contributions are summarized as follows.
\begin{itemize}[leftmargin=*]
    \item We critically examine the limitations of prominent RL-based network models (i.e., DQN from~\citet{Playing_Atari} and DSQN from~\citet{DSQN}) in simultaneous multi-task settings through an experimental case study, thereby highlighting the need for a better solution. 
    \item We introduce the MTSpark methodology, whose key novelty is leveraging bio-inspired spiking operations and architectural enhancements (i.e., active dendrites) for improving multi-task RL quality through task-specific context augmentation.
    Its innovations include the following key points. 
    \begin{itemize}[leftmargin=*]
        \item \textbf{Development of spiking neuron with \textit{active dendrites} (Section~\ref{sec_method_activeD}):} 
        It aims to enhance task specialization and mitigates interference when learning multiple tasks simultaneously.
        Active dendrite is a component that performs modulation for input signals based on the task-specific context, thereby regulating task-specific spiking activities and minimizing task interferences, enabling the model to better distinguish the task and decide the appropriate action to take for improving the overall performance. 
        \item \textbf{Network architecture design (Section~\ref{sec_method_netDesign}):} 
        It targets to accommodate sub-networks specialization for learning individual tasks through the integration of active dendrites. 
        We also propose two variants of active dendrite-equipped spiking architectures: 
        \textit{network with active dendrites} (MTSpark\_AD), and 
        \textit{network with active dendrites \& dueling structure} (MTSpark\_ADD).
        \item \textbf{Multi-task training strategy (Section~\ref{sec_method_training}):} 
        It facilitates the network to effectively learn multiple tasks and minimize task interference by combining the replay mechanism, linear exploration rate, and reward-driven target function. 
    \end{itemize}
    \item We perform a comprehensive evaluation of MTSpark using multiple tasks, i.e., three Atari games (Pong, Breakout, and Enduro).
    In general, MTSpark achieves high scores across three Atari games, having better overall performance than state-of-the-art. 
\end{itemize}

\textbf{Key Results:}
We evaluate the MTSpark using a Python implementation considering three Atari games (i.e., Pong, Breakout, and Enduro), and run it on the Nvidia RTX A6000 multi-GPU machines. 
Remarkably, MTSpark achieves scores of -5.4 for Pong, 0.6 for Breakout, and 371.2 for Enduro, approaching human-level performance (i.e., -3 for Pong, 31 for Breakout, and 368 for Enduro). 
It is also a clear improvement over the state-of-the-art (DQN: -18.6 for Pong, 3.2 for Breakout, and 0 for Enduro; DSQN: -11.2 for Pong, 0.4 for Breakout, and 362.2 for Enduro), incurring similar memory and about $2\times$ lower energy consumption; see labels-\circled{4} and \circled{5} in Figure~\ref{fig_Observation}.

%%%%%%%%%%%%%%%%%%%%%%%%%%%%%%%%%%%%%%%%%%%%%%%%%%%%%%%%%%%%%%%%%%%%%%%%
%%%%%%%%%%%%%%%%%%%%%%%%%%%%%%%%%%%%%%%%%%%%%%%%%%%%%%%%%%%%%%%%%%%%%%%%
\section{Related Works}
\label{sec_relatedworks}

\textbf{Deep Q-Network (DQN):} 
DQN combines Q-learning with DNNs to enable agents to learn optimal policies in high-dimensional environments~\citep{Playing_Atari}. 
In DQN, a neural network (NN) approximates the Q-value function, which predicts the expected future reward for each state-action pair, helping the agent decide the best actions to take. 
DQN stabilizes training by using a replay buffer, storing past experiences and updating the model based on mini-batches of data drawn from this memory, which helps avoid correlations in the data that would hinder learning.
The DQN architecture consists of three convolutional (CONV) layers and two fully connected (FC) layers. 
These CONV layers extract useful features from raw input data (e.g., game frames), while the FC layers compute the Q-values for each possible action in the given state. 

\smallskip
\textbf{Task Interference Mitigation:}
Several methods have been proposed to mitigate task interference and improve learning efficiency across multiple tasks~\citep{Ref_Minhas_NCLsurvey_Access25}, as discussed below. 
\begin{itemize}[leftmargin=*]
    \item \textit{Architectural methods} add task-specific components to the network model, such as extra layers or dedicated modules for each new task, allowing the model to specialize certain parts for specific tasks while sharing common components for generalized learning. 
    However, they face scalability issues as the number of tasks grows due to new layers/modules for each task~\citep{Archi2, PNN}, which are not feasible in large-scale real-world applications.
    \item \textit{Replay methods} periodically reintroduce previously learned tasks during training. 
    They work by either generating synthetic samples or using actual past samples of old tasks, which are then replayed alongside new tasks~\citep{Replay1, Replay2, GenReplay}. 
    However, these methods require revisiting a large number of stored samples to maintain performance. 
    Hence, they are compute- and memory-intensive, demanding significant resources to store and process replayed samples. 
    \item \textit{Regularization methods} constrain the learning process by stabilizing important parameters critical to previous tasks. For instance, Elastic Weight Consolidation (EWC) calculates the significance of each weight for previous tasks and protects these weights during new task learning by adding a regularization term to the loss function, discouraging large changes to important weights~\citep{EWC}, hence retaining knowledge from earlier tasks while adapting to new ones.
\end{itemize}

%%%%%%%%%%%%%%%%%%%%%%%%%%%%%%%%%%%%%%%%%%%%%%%%%%%%%%%%%%%%%%%%%%%%%%%%
%%%%%%%%%%%%%%%%%%%%%%%%%%%%%%%%%%%%%%%%%%%%%%%%%%%%%%%%%%%%%%%%%%%%%%%%
\section{The MTSpark Methodology}
\label{sec_method} 

\begin{figure*}[t]
% \bigskip
    \centering
    \includegraphics[width=\linewidth]{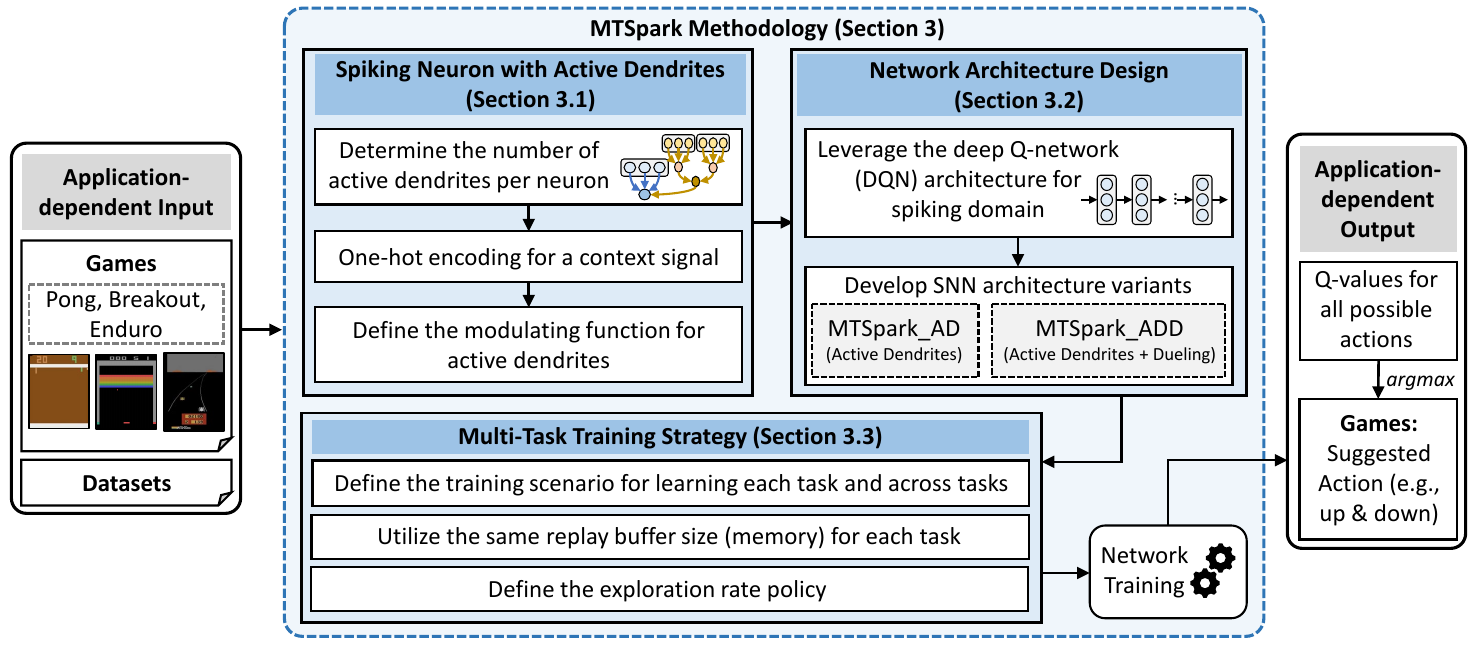}
    \vspace{-0.5cm}
    \caption{The overview of our MTSpark methodology with key steps highlighted in blue: spiking neuron development with active dendrites, network architecture design with two variants (i.e., MTSpark\_AD and MTSpark\_ADD), and multi-task training strategy.} 
    \label{fig_methodology}
\end{figure*}

Our proposed MTSpark methodology addresses energy-efficient simultaneous multi-task RL through network architectural enhancements (i.e., active dendrites) and effective training strategy, as shown in Figure~\ref{fig_methodology} and described in the subsequent sub-sections.

%%%
%%%

%%%%%%%%%%%%%%%%%%%%%%%%%%%%%%%%%%%%%%%%%%%%%%%
\subsection{Spiking Neuron with Active Dendrites}
\label{sec_method_activeD}

\begin{figure}[t]
    \centering
    \includegraphics[width=0.5\linewidth]{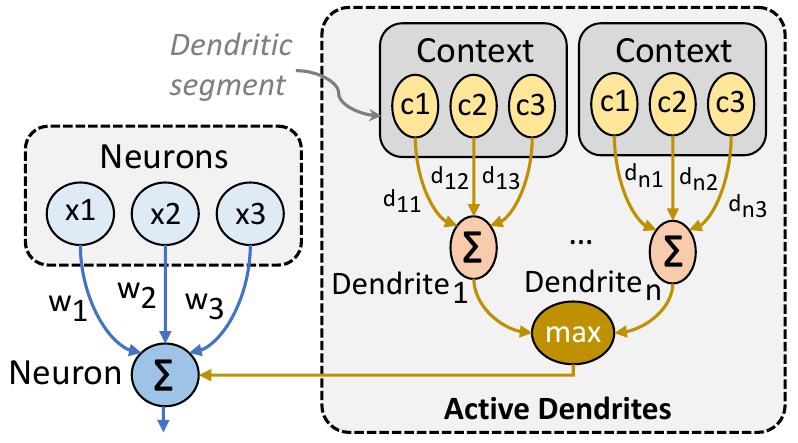}
    \caption{The proposed IF neuron model is enhanced using active dendrites with context signals.}
    \label{fig_ActiveDendrites}
    \bigskip
\end{figure}

This step aims to enhance task specialization and mitigate interference when learning multiple tasks.
To achieve this, \textit{our key novelty in this step is equipping the spiking neurons with active dendrites, which modulate the spiking activity of the integrate-and-fire (IF) neurons based on task-specific context signals}, as shown in Figure~\ref{fig_ActiveDendrites}. 
The context signal ($c$) is provided as input to all of the dendrites on each IF neuron. 
After introducing active dendrites to the IF neurons, the incoming spikes contribute to the membrane potential of the respective neuron, whose behavior can be stated as Equation~\ref{eq_NeuronWithDendrites}.
Here, $d_{j}$ are the dendritic weights associated with the $j^{th}$ dendritic segment, and $f(.)$ is a modulating function.
Meanwhile, $V(t)$ is the neuron's membrane potential at time $t$, and $s_{i}(t)$ is the incoming spike from the $i^{th}$ presynaptic neuron.
If $V(t)$ crosses a predefined threshold ($V_{th}$), the neuron generates a spike.
\begin{equation}
    % \small
    \begin{split}
    V(t) &= V(t - \Delta t) + f\left(\sum_{i}s_{i}(t), \max_{j}(d_j^Tc)\right) \\ \text{with} &\;\;\;\; 
    s(t) = 
    \begin{cases}
        1 & \text{if } V(t) \geq V_{th},\\
        0 & \text{otherwise }
    \end{cases} 
    \end{split}
    \label{eq_NeuronWithDendrites}
\end{equation}

\textbf{Active Dendritic Segment:}
Each IF neuron can be equipped with any number of active dendrites in each dendritic segment. 
To ensure efficient implementation, we utilize the same number of dendrites as the number of tasks. 
To enable task differentiation, we represent each task ($T_{i}$) with a unique one-hot-encoded vector of context signal ($c$). 
This context signal functions as a task identifier, modulating the neuron activations in order to bias the network toward task-relevant pathways and minimize interference across tasks.
We define the modulation function $f(.)$ for active dendrites as Equation~\ref{eq_Modulation}.
The sum of spikes from all presynaptic neurons are weighted by $\sigma{(\max_{j}(d_j^Tc))}$.

\begin{equation}
    \begin{split}
    f\left(\sum_{i}s_{i}(t), \max_{j}(d_j^Tc)\right) = \sum_{i}s_{i}(t) \cdot \sigma{(\max_{j}(d_j^Tc))}
    \end{split}
    \label{eq_Modulation}
\end{equation}

\textbf{Task-Specific Spiking Activity:}
Each spiking IF neuron with active dendrites can selectively adjust its response to a context signal based on the modulation provided by maximal dendritic activation. 
This modulation provides the network with a form of ``gating'' that prevents interference between tasks and minimizes the catastrophic forgetting effects. 
Different tasks exhibit different dendritic activations, and as a result, they contribute differently to the membrane potential of each neuron, leading to different spiking patterns and ``sub-networks'' optimized for each task.

%%%%%%%%%%%%%%%%%%%%%%%%%%%%%%%%%%%%%%%%%
\subsection{Network Architecture Design}
\label{sec_method_netDesign}

This design step aims to accommodate sub-network specialization when learning individual tasks.
To attain this, \textit{our key novelty in this step is leveraging the state-of-the-art DSQN architecture~\citep{DSQN} as the foundation, then enhancing it with active dendrites to develop the MTSpark network architectures}. 
Here, active dendrites have the responsibility of modulating input signals to spiking neurons based on task-specific context, thereby regulating the corresponding spiking activity and establishing task-specific sub-network pathways. 
In this manner, the network model will be able to distinguish the task and select the appropriate sub-network pathways for generating a proper output/action.

\begin{figure}[t]
  \centering
  \includegraphics[width=\linewidth]{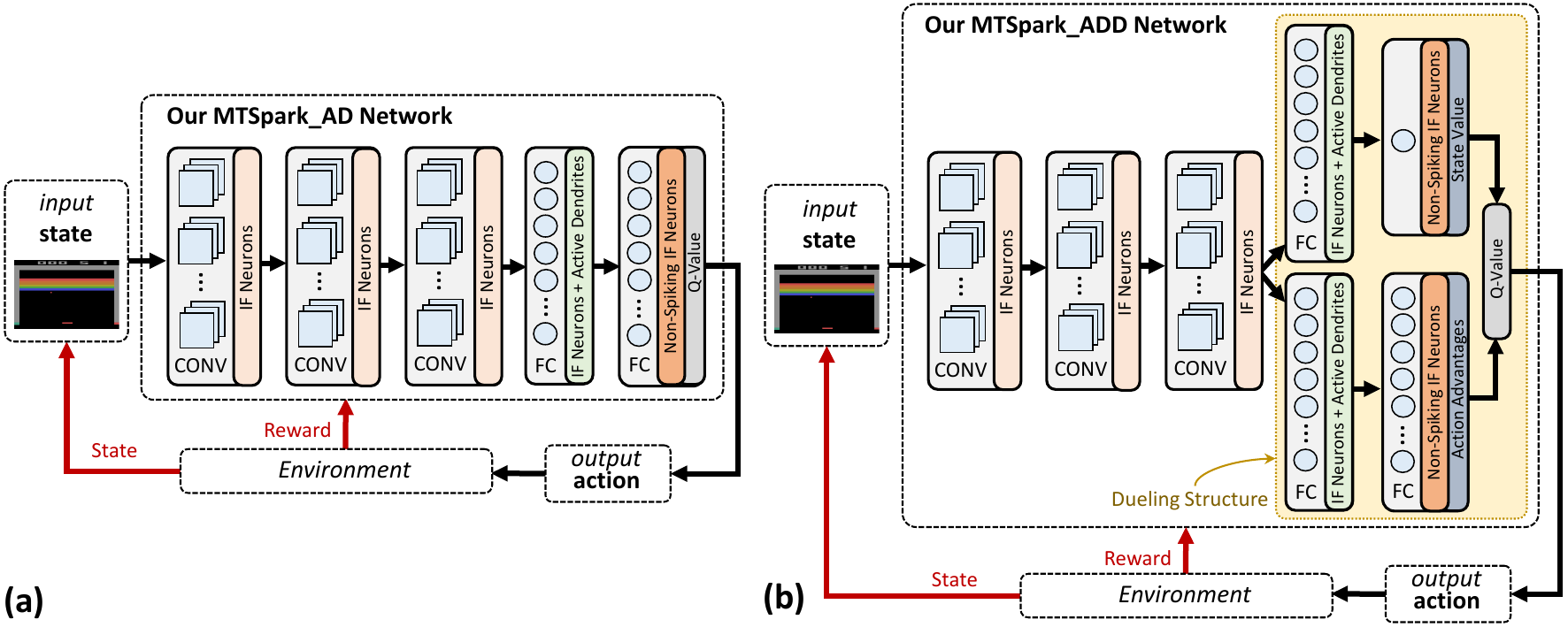}
  \vspace{-0.5cm}
  \caption{Our proposed MTSpark architectures: 
  \textbf{(a)} a network with active dendrites (i.e., MTSpark\_AD), and 
  \textbf{(b)} a network with active dendrites and dueling structure (i.e., MTSpark\_ADD).}
  \label{fig_MTSpark_Archs}
  \bigskip
\end{figure}

\smallskip
\textbf{MTSpark Network Architectures:}
In general, our MTSpark network architecture designs feature three CONV layers with kernel sizes of 8, 4, and 3 and strides of 4, 2, and 1, respectively; each followed by batch normalization and spiking IF neuron layer. 
Input observations are processed through these layers, producing feature maps that are flattened and passed to an FC layer with 512 units. 
Afterward, a spiking neuron layer equipped with active dendrites (AD) and context signals, followed by a final FC layer producing the Q-value over all possible actions in the environment.
To provide trade-offs between performance and network size, we propose two variants of MTSpark network architectures, namely \textit{MTSpark\_AD} and \textit{MTSpark\_ADD}; whose details are described in the following.
\begin{itemize}[leftmargin=*]
    \item \textbf{MTSpark\_AD} employs active dendrites at the FC layer right before the last layer, as shown in Figure~\ref{fig_MTSpark_Archs}(a) and presented in Table~\ref{table_MTSparkAD}.
    This structure targets to provide context signals effectively without incurring significant additional resource overheads by leveraging high-level features that have been extracted by previous CONV layers.
    \item \textbf{MTSpark\_ADD} employs the \textit{dueling structure} on top of the MTSpark\_AD to enhance the performance, thereby forming the MTSpark\_ADD; see Figure~\ref{fig_MTSpark_Archs}(b) and Table~\ref{table_MTSparkADD}. 
    The dueling structure employs two distinct estimators: \textit{the state value function} and \textit{the action-specific advantage function}. 
    This function separation enhances the ability to generalize learning across actions without modifying the RL algorithm~\citep{Ref_Wang_Dueling_ICML16}.
    The state value function gives information regarding how good the current state is regardless of the action, while the advantage function gives information regarding how good the possible actions are, hence helping the agent to improve learning in states where many actions have similar advantages.
    Therefore, MTSpark\_ADD typically provides better performance than MTSpark\_AD at a higher cost (e.g., larger model size).
\end{itemize}

\begin{table}[t]
    \centering
    \begin{minipage}[t]{0.45\linewidth}
        \caption{MTSpark\_AD network architecture.}
          \label{table_MTSparkAD}
          \small
          % \footnotesize
          \centering
            \begin{tabular}{@{}ccc@{}}
            \toprule
            \textbf{Layer} & \textbf{Input} & \textbf{Output} \\ 
             & \textbf{Dimension} & \textbf{Dimension} \\ \midrule
            8$\times$8-Convolution & 84$\times$84$\times$4 & 20$\times$20$\times$32 \\
            BatchNorm & 20$\times$20$\times$32 & 20$\times$20$\times$32 \\
            IF Neuron & 20$\times$20$\times$32 & 20$\times$20$\times$32 \\
            4$\times$4-Convolution & 20$\times$20$\times$32 & 9$\times$9$\times$64 \\
            BatchNorm & 9$\times$9$\times$64 & 9$\times$9$\times$64 \\
            IF Neuron & 9$\times$9$\times$64 & 9$\times$9$\times$64 \\
            3$\times$3-Convolution & 9$\times$9$\times$64 & 7$\times$7$\times$64 \\
            BatchNorm & 7$\times$7$\times$64 & 7$\times$7$\times$64 \\
            IF Neuron & 7$\times$7$\times$64 & 7$\times$7$\times$64 \\ 
            % \midrule
            \multicolumn{3}{c}{----- \textit{Flatten} -----} \\ 
            % \midrule
            FC & 3136$\times$1 & 512$\times$1 \\
            IF Neuron + AD & 512$\times$1 & 512$\times$1 \\
            FC & 512$\times$1 & 18$\times$1 \\
            Non-Spiking IF Neuron & 18$\times$1 & 18$\times$1 \\ \bottomrule
            \end{tabular}
    \end{minipage}
    \hfill
    \begin{minipage}[t]{0.45\linewidth}
        \caption{MTSpark\_ADD network architecture.}
          \label{table_MTSparkADD}
          \small
          % \footnotesize
          \centering
            \begin{tabular}{@{}ccc@{}}
            \toprule
            \textbf{Layer} & \textbf{Input} & \textbf{Output} \\ 
             & \textbf{Dimension} & \textbf{Dimension} \\ \midrule
            8$\times$8-Convolution & 84$\times$84$\times$4 & 20$\times$20$\times$32 \\
            BatchNorm & 20$\times$20$\times$32 & 20$\times$20$\times$32 \\
            IF Neuron & 20$\times$20$\times$32 & 20$\times$20$\times$32 \\
            4$\times$4-Convolution & 20$\times$20$\times$32 & 9$\times$9$\times$64 \\
            BatchNorm & 9$\times$9$\times$64 & 9$\times$9$\times$64 \\
            IF Neuron & 9$\times$9$\times$64 & 9$\times$9$\times$64 \\
            3$\times$3-Convolution & 9$\times$9$\times$64 & 7$\times$7$\times$64 \\
            BatchNorm & 7$\times$7$\times$64 & 7$\times$7$\times$64 \\
            IF Neuron & 7$\times$7$\times$64 & 7$\times$7$\times$64 \\ 
            % \midrule
            \multicolumn{3}{c}{----- \textit{Flatten} -----} \\ 
            % \midrule
            FC & 3136$\times$1 & 512$\times$1 \\
            IF Neuron + AD & 512$\times$1 & 512$\times$1 \\
            \multicolumn{3}{c}{----- \textit{Value} -----} \\
            FC & 512$\times$1 & 1$\times$1 \\
            Non-Spiking Neuron & 1$\times$1 & 1$\times$1 \\
            \multicolumn{3}{c}{----- \textit{Advantage} -----} \\
            FC & 512$\times$1 & 18$\times$1 \\
            Non-Spiking Neuron & 512$\times$1 & 18$\times$1\\ \bottomrule
            \end{tabular}
    \end{minipage}
\end{table}

%%%%%%%%%%%%%%%%%%%%%%%%%%%%%%%%%%
\subsection{Multi-task Training Strategy}
\label{sec_method_training}

This step aims to facilitate the network to effectively learn multiple tasks and minimize task interference. 
To attain this, \textit{our key novelty in this step is combining an efficient replay mechanism, a linear exploration rate, and a reward-driven target function.}
Details of the strategy are described below.
\begin{itemize}[leftmargin=*]
    \item \textbf{Training Scenario and Replay Buffer:}
    We train an MTSpark network (i.e., either MTSpark\_AD or MTSpark\_ADD) on each environment ($E$) for $P$ episodes before switching to the next environment in a round-robin scheduling to enforce the simultaneous multi-task learning scenario. 
    To facilitate this, we maintain a fixed-size replay buffer ($N$) in MTSpark to efficiently store transitions for each environment during the training process.
    \item \textbf{Exploration Rate ($\epsilon$):}
    In line with the widely-used RL training settings~\citep{Playing_Atari, Human_Level_Control}, we linearly anneal the exploration rate $\epsilon$ from an initially high rate to a low rate over the first $T$ frames, and keep it fixed at the low rate thereafter. 
    This allows the agent to freely explore all possible actions in earlier frames and gradually steer the agent toward more effective actions.
    \item \textbf{Target Function:}
    We define our target ($y_n$) as Equation~\ref{eq_NetworkTarget}.
    Here, $r_n$ denotes the reward received after taking an action in the state $\phi_n$, while $\phi_{n + 1}$ denotes the state seen immediately after taking the action, $Q$ denotes our network, and $Q^*$ denotes the target network.
    \begin{equation}
    \begin{split}
        % \small
        & y_n = \begin{cases}
            r_n \;\;\;\;\;\;\;\;\; \text{~if $\phi_{n+1}$ is terminal} \\
            r_n + k \;\;\; \text{~if otherwise}
        \end{cases} \\
        &  \text{with} \;\; k = \gamma Q^{*}(\phi_{n + 1}, \arg\max_{a} Q(\phi_{n + 1}, a)) 
        \label{eq_NetworkTarget}
    \end{split}
    \end{equation}
\end{itemize}

To systematically perform the above techniques, we propose a novel training strategy in Algorithm~\ref{alg:cap}. 

\begin{algorithm}[h]
\caption{Proposed Training Strategy}
\label{alg:cap}
\small
% \footnotesize
% \scriptsize
\begin{algorithmic}[1]
\STATE \textbf{Initialization:}
\STATE \;\;\;\;\; Network $Q$ and target network $Q^*$ with random weights;
\STATE \;\;\;\;\; List of environments $E$;
\STATE \;\;\;\;\; Environment index $i \gets 0$;
\STATE \;\;\;\;\; Replay memory $D$ to capacity $N$ for each environment;
\STATE \;\;\;\;\; Episode counter $\text{episode\_count} \gets 0$;
\STATE \;\;\;\;\; State $\phi$;
\STATE \textbf{Process:}
\FOR{timestep $t = 1$ to $T_{\text{max}}$}
    \STATE With probability $\epsilon$, select a random action $a$; \\ \;\;\;\;\;\; otherwise, $a = \arg \max_{a} Q(\phi, a)$; 
    \STATE Perform action $a$, observe reward $r$ and next state $\phi'$;
    \STATE Store transition $(\phi, a, r, \phi')$ in memory $D_{i}$;
    
    \FOR{environment $j$ in $E$}
        \STATE Sample transitions $(\phi_n, a_n, r_n, \phi_{n+1})$ from $D_{j}$;
        \STATE Set target $y_n$ as defined in Equation \ref{eq_NetworkTarget};
        \STATE Perform gradient descent on $(y_n - Q(\phi_n, a_n))^{2}$;
    \ENDFOR
    
    \IF{$\phi'$ is terminal}
        \STATE Reset state: $\phi \gets$ initial state;
        \STATE $\text{episode\_count} \gets \text{episode\_count} + 1$;
    \ELSE
        \STATE Update state: $\phi \gets \phi'$;
    \ENDIF
    
    \IF{$\text{episode\_count} = 25$}
        \STATE $i \gets (i + 1) \mod \text{number\_of\_environments}$;
        \STATE $\text{episode\_count} \gets 0$;
    \ENDIF
    
    \IF{$t \mod \text{target\_update} = 0$}
        \STATE Update target network: $Q^* \gets Q$;
    \ENDIF
\ENDFOR
\end{algorithmic}
\end{algorithm}
\setlength{\textfloatsep}{1pt}

%%%%%%%%%%%%%%%%%%%%%%%%%%%%%%%%%%%%%%%%%%%%%%%%%%%%%%%%%%%%%%%%%%%%%%%%
\section{Evaluation Methodology}
\label{sec_eval}

%%%%%%%%%%%%
\subsection{Experimental Setup}

We evaluate the MTSpark methodology using Python implementation, and then run it on the Nvidia RTX A6000 48GB multi-GPU machine.
The experimental setup choices (e.g., games/environments and network hyper-parameters) are selected based on the typical settings used in the literature for training an agent with Atari games under RL scenarios~\citealp{Playing_Atari, Human_Level_Control, DSQN}, as described below.

\textbf{Tasks/Environments:} 
We consider three Atari games (i.e., Pong, Breakout, and Enduro) as the targeted tasks (environments), and utilize an action space of 18 possible actions, maintaining consistency in the action space across all environments. 
We focus on a subset of representative Atari games to allow deeper and more controlled evaluation of our methodology, particularly regarding task interference and training dynamics, while providing a range of task diversity and difficulty.
Pong expects a quick response for hitting a ball past the opponent through up/down movements; hence, it has relatively short action-to-reward correlation. 
Breakout expects the agent to destroy all bricks with a ball; hence, it typically has sparse rewards and relatively long action-to-reward correlation. 
Meanwhile, Enduro expects the agent to find a path to overtake other cars while avoiding collisions through right/left movements; hence, it has relatively short action-to-reward correlation.
Furthermore, we also employ 210$\times$160$\times$3 RGB images as inputs, identical to those presented to human players. 

\textbf{Comparison Partners:} 
We consider the prominent DQN~\citep{Playing_Atari} and the state-of-the-art DSQN~\citep{DSQN} as comparison partners. 
For completeness of the study, we also implement their dueling variants, i.e., DQN with dueling structure (DQN\_D) and DSQN with dueling structure (DSQN\_D). 
For all spiking-based networks, we employ timestep $T=4$.

\textbf{Training Settings:}
Details of the hyperparameter settings for the training process in our experiments are summarized in Table~\ref{table_exp_setup}.
All models are trained with the Adam optimizer, a static learning rate of 1e-4, and a batch size of 64 over 4 million frames in each environment. 
Following best practices, we apply a discount factor $\gamma$ of 0.99 to stabilize learning, use a replay buffer of size $2^{20}$ for each environment to store transitions for experience replay, and decay exploration from $\epsilon$=1.00 to $\epsilon$=0.10 over 1 million total frames. 
We collect information on 25 episodes during training in each environment before switching to the next one.

\begin{table}[t]
    \centering
    \caption{Hyperparameter settings for the experiments.}
    \smallskip
    \small
    \centering
    \label{table_exp_setup}
    \begin{tabular}{@{}cc@{}}
    \toprule
    \textbf{Variable}     & \textbf{Value}      \\ \midrule
    Replay Buffer Size ($N$)    & $2^{20}$            \\
    Batch Size            & 64                  \\
    Epsilon Start         & 1.00                \\
    Epsilon End           & 0.10                \\
    Decay Epsilon Over    & 1,000,000 frames    \\
    Update Target Network & Every 10,000 frames \\
    Learning Rate         & 1e-4                \\
    Switch Environments   & Every 25 episodes   \\ \bottomrule
    \end{tabular}
    \bigskip
\end{table}
% \setlength{\textfloatsep}{4pt}

%%%%%%%%%%%%
\subsection{Energy Consumption Model}

To estimate the energy consumption of NN processing, we follow the analytical model proposed in the recent paper from~\citet{Ref_Lemaire_SNNenergyModel_ICONIP22}.
The total energy consumption ($E_{Total}$) is defined as the sum of energy consumption for memory accesses ($E_{Mem}$) and computation ($E_{Com}$), as expressed in Equation~\ref{Eq_Etotal}.
\begin{equation}
    \begin{split}
        E_{Total} = E_{Mem} + E_{Com}
    \end{split}
    \label{Eq_Etotal}
\end{equation}

%%%%
\textbf{Memory Access Energy:}
$E_{Mem}$ in Equation~\ref{Eq_Etotal} can be estimated as the sum of memory read energy ($E_{MemRd}$) and memory write energy ($E_{MemWr}$); see Equation~\ref{Eq_Emem}. 
For DNNs, $E_{MemRd}$ is obtained by multiplying the total number of memory read for input ($Rd^{DNN}_{inp}$) and weight parameters ($Rd^{DNN}_{par}$) with the RAM energy-per-read ($E_{RAM_{Rd}}$), 
while $E_{MemWr}$ is obtained by multiplying the total number of memory write for output ($Wr^{DNN}_{out}$) with the RAM energy-per-write ($E_{RAM_{Wr}}$); see Equation~\ref{Eq_Emem2}.
For SNNs, $E_{MemRd}$ is obtained by multiplying the total number of memory read for input ($Rd^{SNN}_{inp}$), weight parameters ($Rd^{SNN}_{par}$), and membrane potentials ($Rd_{pot}$) with the $E_{RAM_{Rd}}$, 
while $E_{MemWr}$ is obtained by multiplying the total number of memory write for output ($Wr^{SNN}_{out}$) and membrane potentials ($Wr_{pot}$) with the $E_{RAM_{Wr}}$; see Equation~\ref{Eq_Emem2}.
For simplicity, we consider that $E_{RAM_{Rd}} = E_{RAM_{Wr}}$. 
\begin{equation}
    \begin{split}
        E_{Mem} = E_{MemRd} + E_{MemWr}
    \end{split}
    \label{Eq_Emem}
\end{equation}
\begin{equation}
    \begin{split}
        E^{DNN}_{Mem} & = (Rd^{DNN}_{inp} + Rd^{DNN}_{par}) \cdot E_{RAM_{Rd}} + Wr^{DNN}_{out} \cdot E_{RAM_{Wr}} \\
        E^{SNN}_{Mem} & = (Rd^{SNN}_{inp} + Rd^{SNN}_{par} + Rd_{pot}) \cdot E_{RAM_{Rd}} + (Wr^{DNN}_{out} + Wr_{pot}) \cdot E_{RAM_{Wr}}
    \end{split}
    \label{Eq_Emem2}
\end{equation}

To further define the above-mentioned variables, several related operations are considered, including (1) read operations for inputs in DNNs and SNNs, (2) read operations for weight parameters in DNNs and SNNs, (3) read and write operations for membrane parameters in SNNs, and (4) write operations for outputs in DNNs and SNNs; see further details in~\citet{Ref_Lemaire_SNNenergyModel_ICONIP22}.

%%%%%%%%%%%
\textbf{Computation Energy:}
$E_{Com}$ from Equation~\ref{Eq_Etotal} can be estimated as the sum of multiplication-and-accumulation (MAC) energy ($E_{MAC}$) and bias accumulation energy ($E_{ACC}$).
For both DNNs and SNNs, $E_{MAC}$ is obtained by multiplying the energy of addition ($E_{add}$) and multiplication ($E_{mul}$) with the number of MAC ($MAC_{Com}$), while $E_{ACC}$ is obtained by multiplying the $E_{add}$ with the number of accumulation ($ACC_{Com}$); see Equation~\ref{Eq_Ecom}.
\begin{equation}
    \begin{split}
        E_{Com} & = E_{MAC} + E_{ACC} \\ 
        & = (E_{add} + E_{mul}) \cdot MAC_{Com} + E_{add} \cdot ACC_{Com}
    \end{split}
    \label{Eq_Ecom}
\end{equation}
Computation cost encompasses two parts: \textit{elementary operations (ops)}, such as MAC and accumulation; as well as \textit{addressing (addr)} to identify which data to process.
Therefore, $MAC_{Com}$ and $ACC_{Com}$ in Equation~\ref{Eq_Ecom} can be extracted further into Equation~\ref{Eq_Ecom3}. 
Here, $MAC_{ops}$ and $ACC_{ops}$ denote the number of operations for MAC and bias accumulation, respectively.
Meanwhile, $MAC_{addr}$ and $ACC_{addr}$ denote the addressing counts during MAC and bias accumulation, respectively.
\begin{equation}
    \begin{split}
        % \add{MAC^{DNN}_{Com} = MAC^{DNN}_{ops} + MAC^{DNN}_{addr} \;\;\;} & \add{\text{and} \;\;\; ACC^{DNN}_{Com} = ACC^{DNN}_{ops} + ACC^{DNN}_{addr}} \\
        % %
        % \add{MAC^{SNN}_{Com} = MAC^{SNN}_{ops} + MAC^{SNN}_{addr} \;\;\;} & \add{\text{and} \;\;\; ACC^{SNN}_{Com} = ACC^{SNN}_{ops} + ACC^{SNN}_{addr}}
        %
        MAC_{Com} = MAC_{ops} + MAC_{addr} \;\;\; & \text{and} \;\;\; ACC_{Com} = ACC_{ops} + ACC_{addr}
    \end{split}
    \label{Eq_Ecom3}
\end{equation}

To further define the above-mentioned variables, we consider layer-wise computation, comprising components from CONV and FC layers of DNNs and SNNs; see further details in~\citet{Ref_Lemaire_SNNenergyModel_ICONIP22}.

\textbf{Operation-wise Energy Cost:} To calculate the energy consumption, we leverage the energy consumption for each operation (i.e., an addition, a multiplication, and a memory access) shown in Table~\ref{table_energyvalues} under 45nm CMOS technology; based on studies in~\citet{Ref_Jouppi_10Lessons_ISCA21}. 
% Furthermore, we also consider timesteps $T=4$. 
%
\begin{table}[h]
    \caption{Energy consumption values for operations considered in the experiments.}
    \label{table_energyvalues}
    % \smallskip
    \small
    \centering
    \begin{tabular}{cc}
        \toprule
        \textbf{Operation} & \textbf{Energy} \\
        \textbf{[32-bit]} & \textbf{Consumption [$pJ$]} \\
        \midrule
        Addition & 0.1 \\
        Multiplication & 3.1 \\
        % \hline
        % MAC & 3.2 \\
        % \multicolumn{2}{c}{---} \\
        % \multicolumn{2}{c}{\add{SRAM Access}} \\
        SRAM Access: &  \\
          8 kB & 10 \\
          32 kB & 20 \\
          1 MB & 100 \\
        \bottomrule
    \end{tabular}
\end{table}

%%%%%%%%%%%%%%%%%%%%%%%%%%%%%%%%%%%%%%%%%%%%%%%%%%%%%%%%%%%%%%%%%%%%%%%%
\section{Results and Discussion}
\label{sec_results}

%%%%%%%%%%%%%%%%%%%%%%%%%%%%%%%%%%%%%%%%%%
\subsection{Performance on Task Functionality}

\begin{figure*}[t]
    \centering
    {\includegraphics[width=0.95\linewidth]{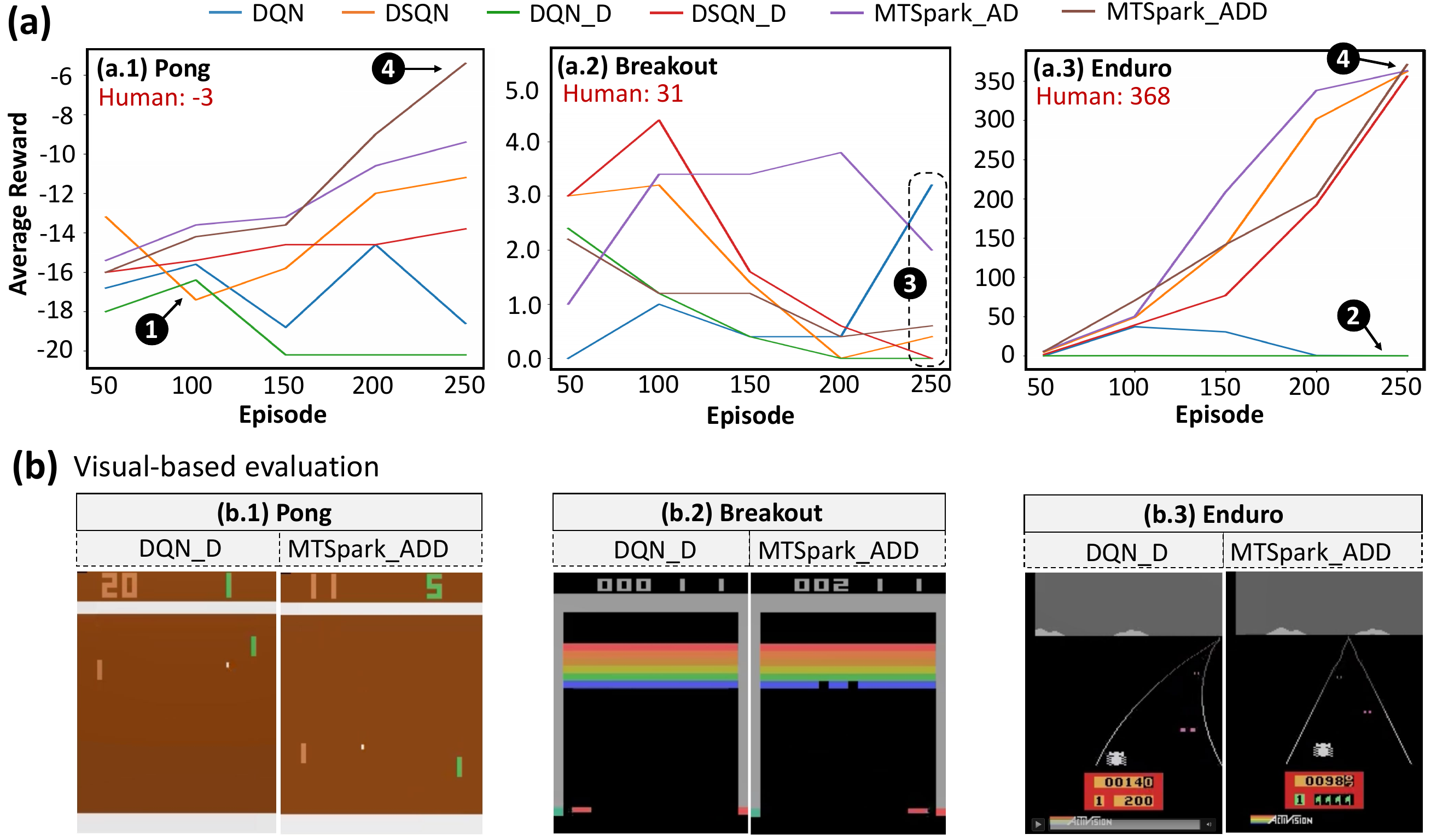}}
    \vspace{-0.2cm}
    \caption{\textbf{(a)} Performance of different models (i.e., DQN, DSQN, DQN\_D, DSQN\_D, MTSpark\_AD, and MTSpark\_ADD) when trained for 250 episodes across three environments/games. 
    \textbf{(b)} Visual-based evaluation of the performance between DQN\_D and MTSpark\_ADD across three Atari games. Game scores show that our MTSpark\_ADD outperforms DQN\_D in all games.}
    \label{fig_results_gamesvisual}
    \bigskip
\end{figure*}

\begin{table*}[t]
\caption{Performance scores of different network models across three environments (i.e., Pong, Breakout, and Enduro) that are obtained from our experiments under the same simultaneous multi-task RL settings. 
Bold text indicates the highest achieved score among all trained models. 
Human scores are identical to those reported in~\citet{Playing_Atari}.
}
\smallskip
\centering
\small
\begin{tabular}{@{}c|cccccc|c@{}}
\toprule
\textbf{Environment} & \textbf{DQN} & \textbf{DSQN} & \textbf{DQN\_D} & \textbf{DSQN\_D} & \textbf{MTSpark\_AD} & \textbf{MTSpark\_ADD} & \textbf{Human} \\ \midrule
\textbf{Pong} & -18.6 & -11.2 & -20.2 & -13.8 & -9.4 & \textbf{-5.4} & -3 \\
\textbf{Breakout} & \textbf{3.2} & 0.4 & 0 & 0 & 2 & 0.6 & 31 \\
\textbf{Enduro} & 0 & 362.2 & 0 & 356 & 363.2 & \textbf{371.2} & 368 \\ \bottomrule
\end{tabular}
\label{table:perf_table}
\bigskip
\end{table*}

Figure~\ref{fig_results_gamesvisual}(a) and Table~\ref{table:perf_table} summarize the performance of various network models across three Atari games: Pong, Enduro, and Breakout. 
The results highlight distinct performance patterns between models with and without spiking components, as well as the competitive advantage of our MTSpark models.
Specifically, network models without spiking components (such as DQN and DQN\_D) exhibit notable performance degradation after certain points during training. 
For instance, their performance drops significantly after point-\circledB{1} in Pong and point-\circledB{2} in Enduro. 
These declines reflect the challenges in managing task interference and encoding long-term action dependencies. 
In contrast, network models with spiking components (such as DSQN, DSQN\_D, and our MTSpark variants) demonstrate more stable performance, as spiking-based operations leverage temporal dynamics for encoding sequential dependencies, thereby mitigating interference between tasks more effectively than non-spiking models. 
Among these, our MTSpark\_AD and MTSpark\_ADD stand out, achieving promising results across all tasks as compared to the comparison partners, and some visual-based evaluations are also presented in Figure~\ref{fig_results_gamesvisual}(b).
Following are the key observations.
\begin{itemize}[leftmargin=*]
    \item In Pong, our MTSpark\_ADD achieves the highest score among all network models with a score of -5.4, approaching human-level performance (with a score of -3). 
    Meanwhile, our MTSpark\_AD also performs strongly with a score of -9.4. 
    In comparison, other spiking models, such as DSQN and DSQN\_D, achieve lower scores, ranging from -11.2 to -20.2, showing their relative struggle to match the adaptive learning efficiency of MTSpark.
    \item In Enduro, MTSpark\_ADD achieves a score of 371.2, slightly surpassing human performance (with a score of 368). 
    MTSpark\_AD follows closely with a score of 363.2. 
    Other spiking models, such as DSQN, show competitive but inferior results, while non-spiking models fall short. 
    This highlights the ability of MTSpark’s active dendrite mechanism and dueling structure to dynamically route task-specific information and adapt during training.
    \item Breakout presents a greater challenge for all models, as shown by \circledB{3}. 
    None of the models achieves scores near human-level performance (with a score of 31). 
    DQN achieves the highest score (with a score of 3.2), followed by MTSpark\_AD (with a score of 2). 
    The challenges posed by Breakout likely stem from its sparse rewards and the need for effective long-term action planning, which may not be adequately addressed by the current architectures.
\end{itemize}
The superior performance of our MTSpark models in Pong and Enduro (see \circledB{4}) can be attributed to their architectural innovations, such as active dendrites and the dueling structure. 
These features allow MTSpark to effectively mitigate task interference by dynamically identifying and routing task-specific sub-networks during the learning process. 
In general, spiking components enhance temporal representation and task-specific adaptation, and the additional enhancements in our MTSpark (i.e., active dendrites and/or dueling structure) give it an edge over other spiking-based network models.

%%%%%%%%%%%%%%%%%%%%%%%%%%%%%%%%%%%%%
\subsection{Network Model Sizes}

Experimental results in Table~\ref{table:arch_params} show that nearly identical parameter counts are observed among DQN, DSQN and MTSpark\_AD models, as well as among DQN\_D, DSQN\_D and MTSpark\_ADD models. 
These results indicate that equipping the IF neurons with active dendrites does not incur significant parameter overhead. 
Moreover, these results even suggest that the improved performance achieved by MTSpark is not derived merely from an increase in model complexity.
Unlike conventional multi-task RL methods that often rely on separate, task-specific network modules or require extensive off-line data storage, MTSpark utilizes a unified parameter set across all tasks, relying only on the context signals to differentiate tasks and exhibit varying network behavior.
MTSpark models also leverage context signals and dendritic modulation to adapt to multiple tasks simultaneously, unlike the approaches that freeze certain parts of the network and/or add task-specific modules or extensive replay. 
These highlight that MTSpark models handle multiple tasks without significantly expanding model sizes. 
% achieving efficiency in both memory usage and computational demands. 

%%%%%%%%%%%%%%%%%%%%%%%%%%%%%%%%%%%%%
\subsection{Energy Consumption}

Energy consumption results are shown in Figure~\ref{fig_energy}. 
In general, spiking networks incur lower energy consumption than their conventional counterparts, mainly due to their less memory accesses and sparse spike-based computation.
For instance, DSQN consumes 34\% less energy than DQN, and DSQN\_D consumes 47\% less energy than DQN\_D. 
We also observe that, MTSpark\_AD has similar energy consumption to DSQN, as they have similar memory access energy and computation, hence leading to about 34\% energy saving from DQN, 66\% energy saving from DQN\_D, and 39\% energy saving from DSQN\_D.
Meanwhile, MTSpark\_ADD has similar energy consumption to DSQN\_D, as they also have similar memory access energy and computation requirements, thereby leading to negligible 5\% energy overhead over DQN and 47\% (about $2\times$) energy saving over DQN\_D. 

\begin{table}[h]
\centering
  \begin{minipage}[h]{0.4\columnwidth}
    \captionof{table}{Number of trainable parameters of different network models, thereby reflecting the model sizes.}
    \label{table:arch_params}
    \small
    \centering
    \begin{tabular}{@{}lc@{}}
      \toprule
      \textbf{Model}  & \textbf{\# Trainable} \\ 
                      & \textbf{Parameters} \\ \midrule
      DQN             & 1,693,682  \\
      DSQN            & 1,693,682  \\
      DQN\_D          & 3,300,339  \\
      DSQN\_D         & 3,300,339  \\
      MTSpark\_AD     & 1,693,691  \\
      MTSpark\_ADD    & 3,300,357  \\ \bottomrule
    \end{tabular}
  \end{minipage}
  \hfill
  \begin{minipage}[h]{0.55\columnwidth}
    \centering
    \includegraphics[width=\linewidth]{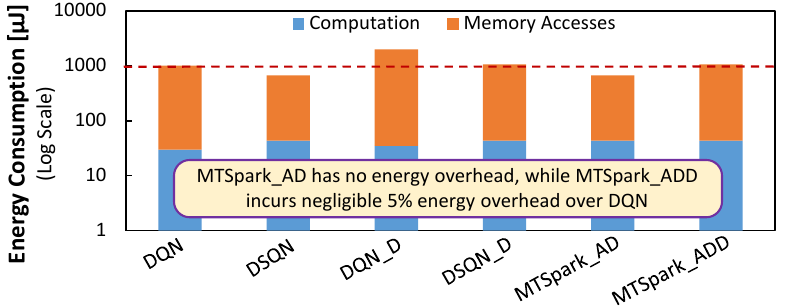}
    \vspace{-0.5cm}
    \captionof{figure}{Energy consumption of different network models: DQN, DSQN, DQN\_D, DSQN\_D, MTSpark\_AD, and MTSpark\_ADD.}
    \label{fig_energy}
  \end{minipage}
\bigskip
\end{table}

%%%%%%%%%%%%%%%%%%%%%%%%%%%%%%%%%%%%%
\subsection{Ablation Study}

%%%%%%%%%
\subsubsection{Impact of Spiking Neurons}
To isolate the effect of spiking neurons, we compare DQN vs. DSQN and DQN\_D vs. DSQN\_D. 
From experimental results in Table~\ref{table:perf_table}, we observe that \textit{spiking-based} DSQN achieves overall better performance than \textit{non-spiking-based} DQN in Pong and Enduro, with no memory overhead and lower energy consumption. 
Similar trends are also observed for DSQN\_D and DQN\_D, as \textit{spiking-based} DSQN\_D also achieves overall better performance than \textit{non-spiking-based} DQN\_D in Pong and Enduro, with no memory overhead and lower energy.
In Pong, DQN scores -18.6 and DSQN scores -11.2, while DQN\_D scores -20.2 and DSQN\_D scores -13.8. 
In Enduro, DQN scores 0 and DSQN scores 362.2, while DQN\_D scores 0 and DSQN\_D scores 356. 
In Breakout, the benefits are less clear as DSQN (0.4) performs slightly worse than DQN (3.2), and DSQN\_D (0) is on par with DQN\_D (0), suggesting that its reward mechanism poses challenges for both conventional and spiking networks.    
Performance improvements achieved by spiking networks are attributed to their capabilities to learn long-term dependence through temporal information across frames. 
Meanwhile, energy-efficiency improvements achieved by spiking networks come from their reduced memory accesses and sparse spike-based operations. 

%%%%%%%%%
\subsubsection{Impact of Dueling Structure}
To evaluate the impact of dueling structure, we compare DQN vs. DQN\_D, DSQN vs. DSQN\_D, and MTSpark\_AD vs. MTSpark\_ADD based on the experimental results in Table~\ref{table:perf_table}.
In Pong, DQN\_D (-20.2) performs slightly worse than DQN (-18.6), and similarly, DSQN\_D (-13.8) performs slightly worse than DSQN (-11.2). 
However, when active dendrites are included, the dueling structure outperforms the non-dueling one (MTSpark\_ADD: -5.4 vs. MTSpark\_AD: -9.4), suggesting that the benefits of dueling architecture may emerge when paired with more sophisticated neuronal dynamics (e.g., active dendrites).
In Enduro, results show mixed effects. 
DSQN (362.2) and DSQN\_D (356) have comparable performance, while MTSpark\_ADD (371.2) slightly outperforms MTSpark\_AD (363.2), implying that the dueling structure can yield some gains when used in conjunction with both spiking neurons and active dendrites.
In Breakout, the dueling structure results in plateaus or decreased performance (DQN: 3.2 vs. DQN\_D: 0, DSQN: 0.4 vs. DSQN\_D: 0, and MTSpark\_AD: 2 vs. MTSpark\_ADD: 0.6), implying that the dueling structure still struggles to contribute positively in Breakout.  
These results indicate that relying solely on the dueling structure is insufficient to master all tasks, as it cannot effectively distinguish different tasks.
However, the addition of spiking neurons and active dendrites on top of the dueling structure helps the model to better perform task identification, thereby improving the overall performance.
Furthermore, the dueling structure increases model size and energy consumption by about $2\times$ as compared to non-dueling model counterparts.

%%%%%%%%%
\subsubsection{Impact of Active Dendrites}
To evaluate the impact of active dendrites, we compare DSQN vs. MTSpark\_AD and DSQN\_D vs. MTSpark\_ADD based on the experimental results in Table~\ref{table:perf_table}. 
In general, MTSpark\_AD improves upon DSQN and MTSpark\_ADD improves upon DSQN\_D.
In Pong, MTSpark\_AD scores -9.4 and DSQN scores -11.2, while MTSpark\_ADD scores -5.4 and DSQN\_D scores -13.8. 
In Breakout, MTSpark\_AD scores 2 and DSQN scores 0.4, while MTSpark\_ADD scores 0.6 and DSQN\_D scores 0. 
In Enduro, MTSpark\_AD scores 363.2 and DSQN scores 362.2, while MTSpark\_ADD scores 371.2 and DSQN\_D scores 356. 
These results suggest a consistent positive impact of active dendrites in defining effective sub-network pathways for targeted tasks, thereby leading to better task-specific responses and better long-term planning for learning tasks with complex input or delayed rewards.

%%%%%%%%%%%%%%%%%%%%%%%%%%%%%%%%%%%%%
\subsection{Applicability of MTSpark for Different Multi-Task Applications}

To evaluate the versatility of MTSpark for different multi-task applications, we assess its performance on image classification tasks based on different datasets.
In multi-task image classification, one task is represented with a specific dataset. 
Here, we consider MNIST~\citep{MNIST}, Fashion MNIST~\citep{FMNIST}, CIFAR-10~\citep{krizhevsky2014cifar}, and Imagenette (10-class subset of ImageNet) datasets since they have been widely used for evaluating SNNs~\citep{Ref_Diehl_SNN_FNCOM15, Ref_Putra_FSpiNN_TCAD20, Ref_Putra_TopSpark_IROS23}, and they have the same number of output space (i.e., 10 classes). 
\begin{itemize}[leftmargin=*]
    \item MNIST represents a simple dataset, which has 60,000 training images and 10,000 testing images, each having a dimension of 28x28 pixels and 1 channel (grayscale).
    \item Fashion MNIST also represents a simple dataset, that has 60,000 training images and 10,000 testing images, each having a dimension of 28x28 pixels and 1 channel (grayscale).
    \item CIFAR-10 represents a medium complexity dataset, that has 50,000 training images and 10,000 testing images, each having 32×32 pixels and 3 channels (RGB).
    \item Imagenette represents a more complex dataset with 10-class subset of ImageNet, that has 9,469 training images and 3,925 testing images, each having 160×160 pixels and 3 channels (RGB).
\end{itemize}

%%%%
\textbf{Network Architectures:}
We compare MTSpark against: (1) a baseline DNN, and (2) an EWC-based DNN, since EWC is a prominent and widely-used technique for mitigating task interference~\citep{Ref_Minhas_NCLsurvey_Access25}. 
The baseline DNN architecture is compact and tailored for standard image classification tasks. 
It comprises three CONV layers with 32, 64, and 64 filters, using filter sizes of 8$\times$8, 4$\times$4, and 3$\times$3, respectively, and followed by batch normalization and ReLU activation. 
The extracted features are then flattened and passed through an FC layer with 128 units before mapping to the 10 output classes.
MTSpark adopts the same architecture but replaces the ReLU activations with IF neurons equipped with active dendrites.

\begin{figure}[h]
    \centering
    {\includegraphics[width=0.9\linewidth]{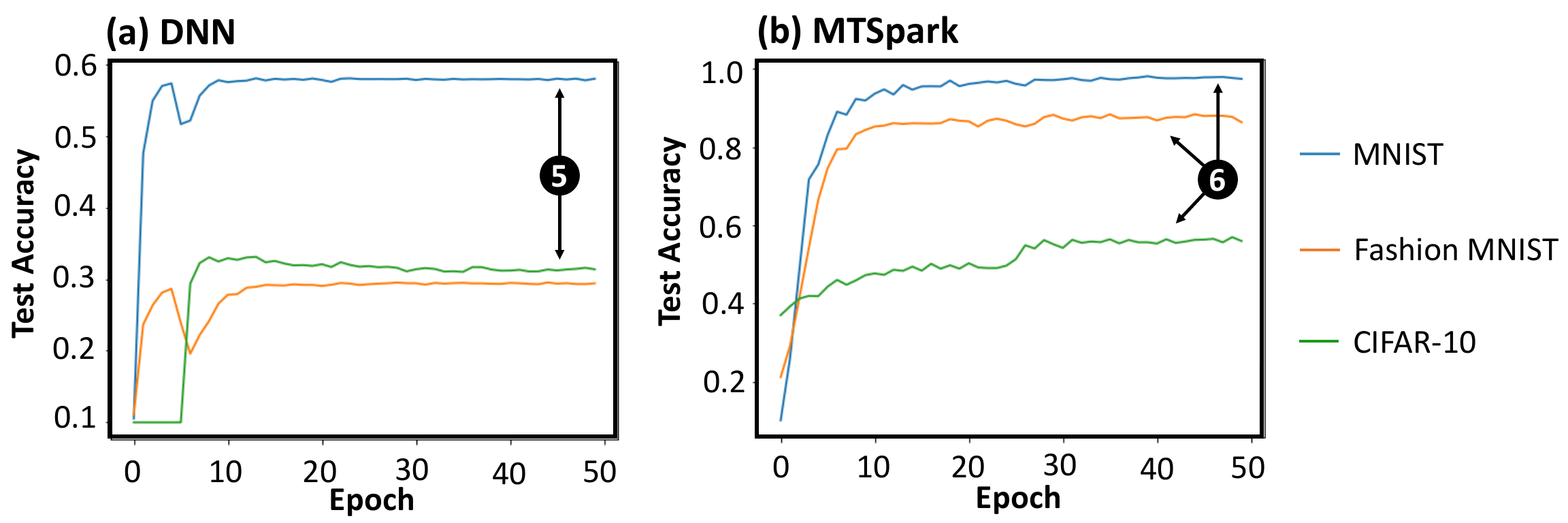}}
    \caption{Test accuracy of different network models: \textbf{(a)} DNN and \textbf{(b)} our MTSpark, on three different image classification tasks: MNIST, Fashion MNIST, and CIFAR-10.}
    \label{fig:comp_class}
    \bigskip
\end{figure}

%%%%
\textbf{Comparison with the Baseline DNN:}
Figure~\ref{fig:comp_class} shows the experimental results after performing simultaneous multi-task learning, i.e., the network is trained with different tasks/datasets at once.
For MNIST, the baseline DNN achieves only modest performance, plateauing at around 57\% accuracy after 50 epochs; see the blue line of \circledB{5}. 
In contrast, MTSpark demonstrates a steep learning curve, surpassing 92\% accuracy by epoch 10 and achieving approximately 98\% by the end of training; see the blue line of \circledB{6}.
A similar trend is observed for Fashion MNIST. 
The baseline DNN performs poorly, reaching only 29\% accuracy by the final epoch; see the orange line of \circledB{5}. 
Meanwhile, MTSpark rapidly achieves 84\% accuracy within the first 10 epochs and concludes training with an accuracy of 88\%; see the orange line of \circledB{6}.
Performance on CIFAR-10, which is relatively more complicated than both MNIST and Fashion MNIST, also follows similar trends. 
While the baseline DNN struggles, achieving a maximum accuracy of only 32\%, MTSpark begins with an initial accuracy of 37\% and steadily improves to approximately 57\% by the end of training; see green line of \circledB{6}. 
These results underscore MTSpark's superiority across diverse image classification tasks, as its ability to leverage active dendrites for task-specific context processing enables faster, more effective learning than the conventional DNN.

%%%%
\textbf{Comparison with the EWC-based DNN:}
To benchmark our MTSpark against the EWC-based DNN, we subsequently train the same baseline DNN on MNIST, Fashion MNIST, and CIFAR-10. 
EWC is applied to the DNN with four different regularization weights (i.e., 1e2, 1e4, 1e6, and 1e8) to assess its ability to balance knowledge retention and adaptability. 
The experimental results are summarized in Table~\ref{tab_ewc_mt_comp}.
These results show that, the EWC-based DNN struggles to deliver competitive performance across all datasets. 
Even at its best, accuracy remains low, with values of 33.5\% for MNIST, 48.2\% for Fashion MNIST, and 21.7\% for CIFAR-10. 
This highlights the challenges posed by multi-task learning. 
In contrast, our MTSpark achieves significantly higher accuracies: 97.5\% for MNIST, 86.4\% for Fashion MNIST, and 56.0\% for CIFAR-10. 
This performance illustrates the ability of MTSpark to learn and retain knowledge across tasks more effectively than EWC.

\begin{table}[h]
\caption{Performance comparison between the EWC-based DNN and the MTSpark-based model at the end of training on image classification tasks.}
\smallskip
\small
\centering
\begin{tabular}{@{}cccc@{}}
\toprule
\textbf{Model} & \textbf{MNIST} & \textbf{Fashion MNIST} & \textbf{CIFAR-10} \\ \midrule
EWC\_1e2 & 10.5\% & 13.8\% & 20.3\% \\
EWC\_1e4 & 12.1\% & 16.8\% & 21.7\% \\
EWC\_1e6 & 33.5\% & 39.4\% & 18.0\% \\
EWC\_1e8 & 30.0\% & 48.2\% & 12.8\% \\
MTSpark  & 97.5\% & 86.4\% & 56.0\% \\ \bottomrule
\end{tabular}
\label{tab_ewc_mt_comp}
\bigskip
\end{table}

%%%%
\textbf{Different Multi-Task Scenarios:}
We also conduct an ablation study to evaluate the impact of conventional DNN and MTSpark architectures on various multi-task learning scenarios. 
Specifically, we perform experimental case studies for two scenarios: (1) \textit{two-task learning}, using the MNIST and Fashion MNIST datasets; and (2) \textit{three-task learning}, using the MNIST, Fashion MNIST, and Imagenette (10-class subset of ImageNet) datasets. 
Experimental results are provided in Figure~\ref{fig_comp_class_2task} and Figure~\ref{fig_comp_class_3task} for two-task and three-task learning, respectively.
Following are the key observations.
\begin{itemize}[leftmargin=*]
    \item \textit{Two-task learning scenario}: 
    The DNN achieves high accuracy on a single task (MNIST), but struggles with the second task (Fashion MNIST), as shown in Figure~\ref{fig_comp_class_2task}(a). 
    This is attributed to the DNN architecture, which is optimized for learning specific tasks, and consequently, it exhibits frequent fluctuations in its learning curve when exposed to a new task.
    Conversely, MTSpark-based model demonstrates superior performance by achieving high accuracy on both tasks. 
    Its learning curves are smoother and converge earlier in the training process, as shown in Figure~\ref{fig_comp_class_2task}(b). 
    This performance improvement is due to MTSpark’s active dendrite mechanisms, which effectively modulate spiking activity based on contextual signals, enabling efficient multi-task learning.
    \begin{figure}[t]
    \centering
    {\includegraphics[width=0.9\linewidth]{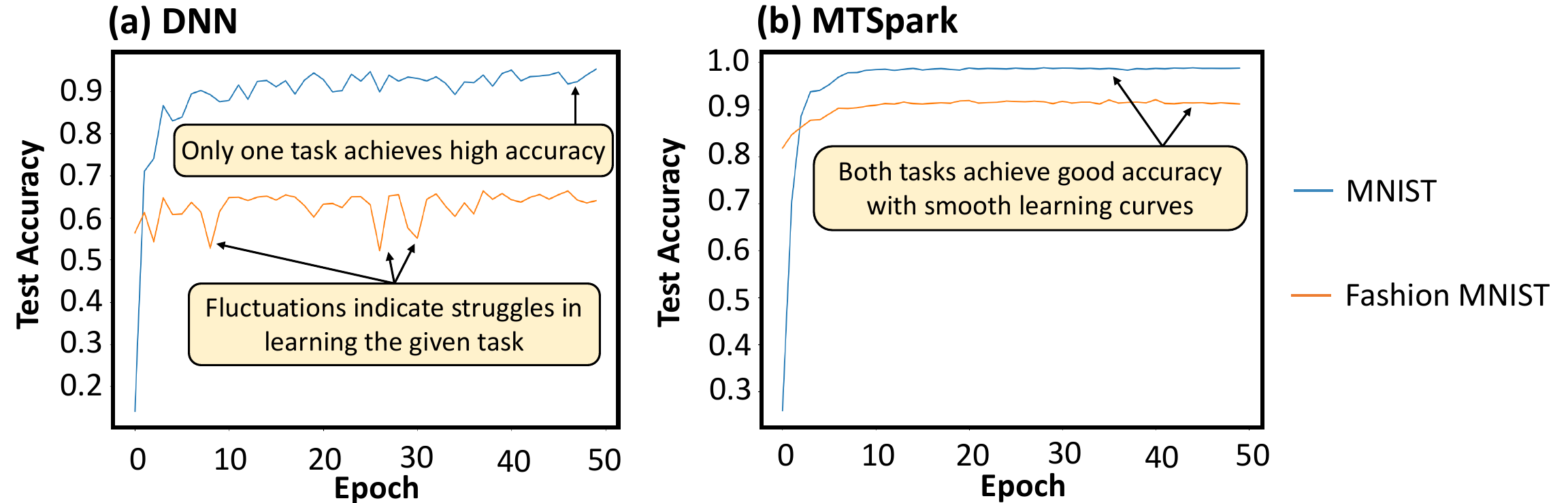}}
    \caption{Test accuracy in the two-class learning scenario for \textbf{(a)} DNN and \textbf{(b)} our MTSpark, considering two tasks (i.e., MNIST and Fashion MNIST).}
    \label{fig_comp_class_2task}
    \end{figure}
    \item \textit{Three-task learning scenario}: 
    In this case, we observe that the limitations of the DNN are further amplified. 
    The DNN achieves high accuracy on MNIST, moderate accuracy on Fashion MNIST, and very low accuracy on Imagenette, as shown Figure~\ref{fig_comp_class_3task}(a). 
    Adding additional tasks leads to significant accuracy degradation on previously learned tasks and a failure to generalize to new ones, reflecting inherent limitations of the DNN in multi-task learning. 
    In contrast, MTSpark-based model achieves high accuracy on MNIST and Fashion MNIST while delivering better accuracy on Imagenette compared to the DNN. 
    Its learning curves remain smooth, as shown in Figure~\ref{fig_comp_class_3task}(b), highlighting its ability to learn multiple tasks more effectively due to its active dendrites.
    \begin{figure}[t]
    \centering
    {\includegraphics[width=0.9\linewidth]{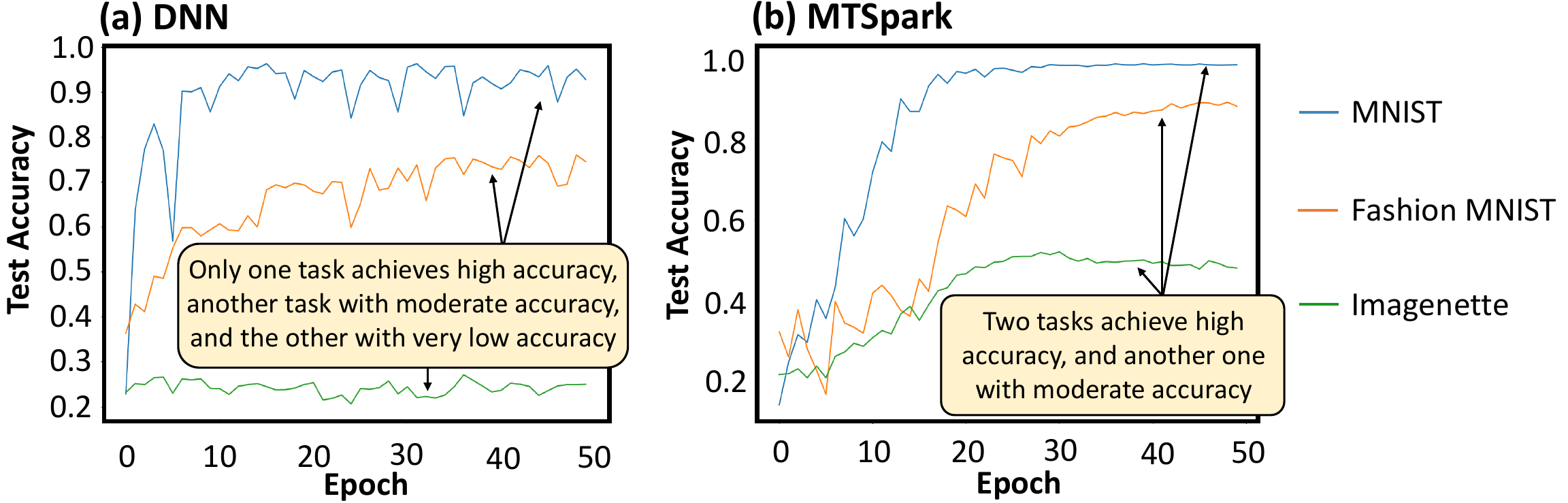}}
    \caption{Test accuracy in the three-class learning scenario \textbf{(a)} DNN and \textbf{(b)} our MTSPark, considering three tasks (i.e., MNIST, Fashion MNIST, and Imagenette).}
    \label{fig_comp_class_3task}
    \bigskip
    \end{figure}
\end{itemize}

%%%%%%%%%%%%%%%%%%%%%%%%%%%%%%%%%%%%%%%%%%%%%%%%%%%%%%%%%%%%%%%%%%%%%%%%%%%%%
%%%%%%%%%%%%%%%%%%%%%%%%%%%%%%%%%%%%%%%%%%%%%%%%%%%%%%%%%%%%%%%%%%%%%%%%%%%%%
\section{Conclusion}
\label{Sec_Conclude}

We present a novel MTSpark methodology to enables simultaneous multi-task RL using bio-inspired spiking networks. 
MTSpark exploits sparse spike-based operations to achieve energy-efficient computation, and leverages temporal information within spike trains to learn long-term dependencies.
MTSpark leverages active dendrites, dueling structure, and task-specific context signals for enabling dynamic and context-sensitive task differentiation, thereby regulating task-specific spiking activities and minimizing task interferences. 
As a result, MTSpark models can better distinguish the task and decide the appropriate action to take for improving the overall performance. 
Experimental results demonstrate the effectiveness of MTSpark (i.e., Pong: -5.4, Breakout: 0.6, and Enduro: 371.2), surpassing state-of-the-art, while incurring similar memory footprint and $2\times$ lower energy consumption. 
These results also show the potential of our MTSpark in advancing multi-task RL toward enabling energy-efficient bio-inspired generalist agents.

%%%%%%%%%%%%%%%%%%%%%%%%%%%%%%%%%%%%%%%%%%%%%%%%%%%%%%%%%%%%%%%%%%%%%%%%%%%%%%%%%%%%%%%%%%%%%
%%%%%%%%%%%%%%%%%%%%%%%%%%%%%%%%%%%%%%%%%%%%%%%%%%%%%%%%%%%%%%%%%%%%%%%%%%%%%%%%%%%%%%%%%%%%%

% Only add this information once your submission is accepted and deanonymized. 

%%%%%%%%%%%%%%%%%%%%%%%%%%%%%%%%%%%%%%%%%%%%%%%%%%%%%%%%%%%%%%%%%%%%%%%%%%%%%%%%%%%%%%%%%%%%%
%%%%%%%%%%%%%%%%%%%%%%%%%%%%%%%%%%%%%%%%%%%%%%%%%%%%%%%%%%%%%%%%%%%%%%%%%%%%%%%%%%%%%%%%%%%%%

\bibliography{main}
\bibliographystyle{tmlr}

%%%%%%%%%%%%%%%%%%%%%%%%%%%%%%%%%%%%%%%%%%%%%%%%%%%%%%%%%%%%%%%%%%%%%%%%%%%%%%%%%%%%%%%%%%%%%
%%%%%%%%%%%%%%%%%%%%%%%%%%%%%%%%%%%%%%%%%%%%%%%%%%%%%%%%%%%%%%%%%%%%%%%%%%%%%%%%%%%%%%%%%%%%%

% \appendix
% \section{Appendix}

\end{document}

%% file: math_commands.tex
\usepackage{amsmath,amsfonts,bm}

\def\eqref#1{equation~\ref{#1}}
\def\1{\bm{1}}

\DeclareMathAlphabet{\mathsfit}{\encodingdefault}{\sfdefault}{m}{sl}
\SetMathAlphabet{\mathsfit}{bold}{\encodingdefault}{\sfdefault}{bx}{n}

%% file: main.bib
@ARTICLE{Ref_Putra_SpikeNAS_TAI26,
  author={Putra, Rachmad Vidya Wicaksana and Shafique, Muhammad},
  journal={IEEE Transactions on Artificial Intelligence}, 
  title={SpikeNAS: A Fast Memory-Aware Neural Architecture Search Framework for Spiking Neural Network-Based Embedded AI Systems}, 
  year={2026},
  volume={7},
  number={2},
  pages={947-959},
  doi={10.1109/TAI.2025.3586238}}

@inproceedings{Ref_Cachi_TMSNN_IWANN23,
  title={TM-SNN: threshold modulated spiking neural network for multi-task learning},
  author={Cachi, Paolo G and Soto, Sebasti{\'a}n Ventura and Cios, Krzysztof J},
  booktitle={International Work-Conference on Artificial Neural Networks},
  pages={653--663},
  year={2023},
  organization={Springer}
}

@inproceedings{Ref_Hessel_PopArt_AAAI19,
  title={Multi-task deep reinforcement learning with popart},
  author={Hessel, Matteo and Soyer, Hubert and Espeholt, Lasse and Czarnecki, Wojciech and Schmitt, Simon and Van Hasselt, Hado},
  booktitle={Proceedings of the AAAI Conference on Artificial Intelligence},
  volume={33},
  number={01},
  pages={3796--3803},
  year={2019}
}

@inproceedings{Ref_Espeholt_IMPALA_ICML18,
  title={Impala: Scalable distributed deep-rl with importance weighted actor-learner architectures},
  author={Espeholt, Lasse and Soyer, Hubert and Munos, Remi and Simonyan, Karen and Mnih, Vlad and Ward, Tom and Doron, Yotam and Firoiu, Vlad and Harley, Tim and Dunning, Iain and others},
  booktitle={International conference on machine learning},
  pages={1407--1416},
  year={2018},
  organization={PMLR}
}

@article{DEramo2020SharingKIA,
  title={Sharing Knowledge in Multi-Task Deep Reinforcement Learning},
  author={Carlo D'Eramo and Davide Tateo and Andrea Bonarini and Marcello Restelli and J. Peters},
  journal={ArXiv},
  year={2020},
  volume={abs/2401.09561},
  url={https://api.semanticscholar.org/CorpusId:209479151}
}

@inproceedings{Sodhani2021MultiTaskRLA,
  title={Multi-Task Reinforcement Learning with Context-based Representations},
  author={Shagun Sodhani and Amy Zhang and Joelle Pineau},
  booktitle={International Conference on Machine Learning},
  year={2021},
  url={https://api.semanticscholar.org/CorpusId:231879645}
}

@article{Zhang2018AnOOA,
  title={An Overview of Multi-task Learning},
  author={Yu Zhang and Qiang Yang},
  journal={National Science Review},
  year={2018},
  volume={5},
  pages={30-43},
}

@INPROCEEDINGS{Ref_Minhas_Replay4NCL_DAC25,
  author={Minhas, Mishal Fatima and Putra, Rachmad Vidya Wicaksana and Awwad, Falah and Hasan, Osman and Shafique, Muhammad},
  booktitle={2025 62nd ACM/IEEE Design Automation Conference (DAC)}, 
  title={Replay4NCL: An Efficient Memory Replay-based Methodology for Neuromorphic Continual Learning in Embedded AI Systems}, 
  year={2025},
  volume={},
  number={},
  pages={1-7},
  doi={10.1109/DAC63849.2025.11132839}}

@ARTICLE{Ref_Minhas_NCLsurvey_Access25,
  author={Minhas, Mishal Fatima and Putra, Rachmad Vidya Wicaksana and Awwad, Falah and Hasan, Osman and Shafique, Muhammad},
  journal={IEEE Access}, 
  title={Continual Learning With Neuromorphic Computing: Foundations, Methods, and Emerging Applications}, 
  year={2025},
  volume={13},
  number={},
  pages={124824-124873},
  doi={10.1109/ACCESS.2025.3588665}}

@INPROCEEDINGS{Ref_Jouppi_10Lessons_ISCA21,
  author={Jouppi, Norman P. and Hyun Yoon, Doe and Ashcraft, Matthew and Gottscho, Mark and Jablin, Thomas B. and Kurian, George and Laudon, James and Li, Sheng and Ma, Peter and Ma, Xiaoyu and Norrie, Thomas and Patil, Nishant and Prasad, Sushma and Young, Cliff and Zhou, Zongwei and Patterson, David},
  booktitle={2021 ACM/IEEE 48th Annual International Symposium on Computer Architecture (ISCA)}, 
  title={Ten Lessons From Three Generations Shaped Google’s TPUv4i : Industrial Product}, 
  year={2021},
  volume={},
  number={},
  pages={1-14},
  doi={10.1109/ISCA52012.2021.00010}}

@inproceedings{Ref_Lemaire_SNNenergyModel_ICONIP22,
  title={An analytical estimation of spiking neural networks energy efficiency},
  author={Lemaire, Edgar and Cordone, Lo{\"\i}c and Castagnetti, Andrea and Novac, Pierre-Emmanuel and Courtois, Jonathan and Miramond, Beno{\^\i}t},
  booktitle={International Conference on Neural Information Processing (ICONIP)},
  pages={574--587},
  year={2022},
  organization={Springer}
}

@article{Mastering_Chess_Shogi,
  title={Mastering chess and shogi by self-play with a general reinforcement learning algorithm},
  author={Silver, David and Hubert, Thomas and Schrittwieser, Julian and Antonoglou, Ioannis and Lai, Matthew and Guez, Arthur and Lanctot, Marc and Sifre, Laurent and Kumaran, Dharshan and Graepel, Thore and others},
  journal={arXiv preprint arXiv:1712.01815},
  year={2017}
}

@article{Playing_Atari,
  title={Playing atari with deep reinforcement learning},
  author={Mnih, Volodymyr and Kavukcuoglu, Koray and Silver, David and Graves, Alex and Antonoglou, Ioannis and Wierstra, Daan and Riedmiller, Martin},
  journal={arXiv preprint arXiv:1312.5602},
  year={2013}
}

@article{Human_Level_Control,
  title={Human-level control through deep reinforcement learning},
  author={Mnih, Volodymyr and Kavukcuoglu, Koray and Silver, David and Rusu, Andrei A and Veness, Joel and Bellemare, Marc G and Graves, Alex and Riedmiller, Martin and Fidjeland, Andreas K and Ostrovski, Georg and others},
  journal={Nature},
  volume={518},
  number={7540},
  pages={529--533},
  year={2015},
  publisher={Nature Publishing Group}
}

@inproceedings{Replay1,
  title={Learning to Learn without Forgetting by Maximizing Transfer and Minimizing Interference},
  author={Riemer, Matthew and Cases, Ignacio and Ajemian, Robert and Liu, Miao and Rish, Irina and Tu, Yuhai and Tesauro, Gerald},
  booktitle={International Conference on Learning Representations (ICLR)}
}

@article{Replay2,
  title={Gradient episodic memory for continual learning},
  author={Lopez-Paz, David and Ranzato, Marc'Aurelio},
  journal={Advances in Neural Information Processing Systems (NIPS)},
  volume={30},
  year={2017}
}

@inproceedings{Active_Image,
  title={Active Dendrites Enable Efficient Continual Learning in Time-To-First-Spike Neural Networks},
  author={Pes, Lorenzo and Luiken, Rick and Corradi, Federico and Frenkel, Charlotte},
  booktitle={IEEE 6th International Conference on AI Circuits and Systems (AICAS)},
  pages={41--45},
  year={2024},
  organization={IEEE}
}

@article{Archi2,
  author       = {Jeongtae Lee and others},
  title        = {Lifelong Learning with Dynamically Expandable Networks},
  journal      = {CoRR},
  volume       = {abs/1708.01547},
  year         = {2017},
  url          = {http://arxiv.org/abs/1708.01547},
  eprinttype    = {arXiv},
  eprint       = {1708.01547},
  bibsource    = {dblp computer science bibliography, https://dblp.org}
}

@article{DSQN,
  title={Deep reinforcement learning with spiking q-learning},
  author={Chen, Ding and Peng, Peixi and Huang, Tiejun and Tian, Yonghong},
  journal={arXiv preprint arXiv:2201.09754},
  year={2022}
}

@article{EWC,
  title={Overcoming catastrophic forgetting in neural networks},
  author={Kirkpatrick, James and others},
  journal={Proceedings of the National Academy of Sciences (PNAS)},
  volume={114},
  number={13},
  pages={3521--3526},
  year={2017},
  publisher={National Acad Sciences}
}

@article{PNN,
  author       = {Andrei A. Rusu and others},
  title        = {Progressive Neural Networks},
  journal      = {CoRR},
  volume       = {abs/1606.04671},
  year         = {2016},
  url          = {http://arxiv.org/abs/1606.04671},
  eprinttype    = {arXiv},
  eprint       = {1606.04671},
  bibsource    = {dblp computer science bibliography, https://dblp.org}
}

@article{GenReplay,
  title={Continual learning with deep generative replay},
  author={Shin, Hanul and others},
  journal={Advances in Neural Information Processing Systems (NIPS)},
  volume={30},
  year={2017}
}

@INPROCEEDINGS{Ref_Putra_TopSpark_IROS23,
  author={Putra, Rachmad Vidya Wicaksana and Shafique, Muhammad},
  booktitle={2023 IEEE/RSJ International Conference on Intelligent Robots and Systems (IROS)}, 
  title={TopSpark: A Timestep Optimization Methodology for Energy-Efficient Spiking Neural Networks on Autonomous Mobile Agents}, 
  year={2023},
  volume={},
  number={},
  pages={3561-3567},
  doi={10.1109/IROS55552.2023.10342499}}

@ARTICLE{Ref_Diehl_SNN_FNCOM15,
AUTHOR={Diehl, Peter and Cook, Matthew},   
TITLE={Unsupervised learning of digit recognition using spike-timing-dependent plasticity},      
JOURNAL={Frontiers in Computational Neuroscience},      
VOLUME={9},      
PAGES={99},     
YEAR={2015},      
URL={https://www.frontiersin.org/article/10.3389/fncom.2015.00099},     
DOI={10.3389/fncom.2015.00099},      
ISSN={1662-5188},   
}

@ARTICLE{Ref_Putra_FSpiNN_TCAD20,
  author={Rachmad Vidya Wicaksana {Putra} and Muhammad {Shafique}},
  journal={IEEE Transactions on Computer-Aided Design of Integrated Circuits and Systems (TCAD)}, 
  title={\uppercase{FS}pi\uppercase{NN}: \uppercase{A}n Optimization Framework for Memory-Efficient and Energy-Efficient Spiking Neural Networks}, 
  year={2020},
  volume={39},
  number={11},
  pages={3601-3613},
  doi={10.1109/TCAD.2020.3013049}}

@inproceedings{Ref_SpikeDyn,
  title={Spikedyn: A framework for energy-efficient spiking neural networks with continual and unsupervised learning capabilities in dynamic environments},
  author={Putra, Rachmad Vidya Wicaksana and Shafique, Muhammad},
  booktitle={2021 58th ACM/IEEE Design Automation Conference (DAC)},
  pages={1057--1062},
  year={2021},
  organization={IEEE}
}

@inproceedings{Ref_lpSpikeCon,
  title={lpspikecon: Enabling low-precision spiking neural network processing for efficient unsupervised continual learning on autonomous agents},
  author={Putra, Rachmad Vidya Wicaksana and Shafique, Muhammad},
  booktitle={2022 International Joint Conference on Neural Networks (IJCNN)},
  pages={1--8},
  year={2022},
  organization={IEEE}
}

@article{Ref_Bartolozzi_EmbodiedNeuroIntel_Nature22,
  title={Embodied neuromorphic intelligence},
  author={Bartolozzi, Chiara and Indiveri, Giacomo and Donati, Elisa},
  journal={Nature communications},
  volume={13},
  number={1},
  pages={1024},
  year={2022},
  publisher={Nature Publishing Group UK London}
}

@ARTICLE{Ref_Luo_EEGSNN_Access20,
  author={Luo, Yuling and Fu, Qiang and Xie, Juntao and Qin, Yunbai and Wu, Guopei and Liu, Junxiu and Jiang, Frank and Cao, Yi and Ding, Xuemei},
  journal={IEEE Access}, 
  title={EEG-Based Emotion Classification Using Spiking Neural Networks}, 
  year={2020},
  volume={8},
  number={},
  pages={46007-46016},
  doi={10.1109/ACCESS.2020.2978163}
  }

@InProceedings{Ref_Cordone_ObjDetSNN_IJCNN22,
    author    = {Cordone, Loic and Miramond, Benoît and Thierion, Phillipe},
    title     = {Object Detection with Spiking Neural Networks on Automotive Event Data},
    booktitle = {International Joint Conference on Neural Networks (IJCNN)},
    month     = {July},
    year      = {2022},
    pages     = {1--8},
}

@inproceedings{mt_rl,
  title={Multi-task reinforcement learning with context-based representations},
  author = {Sodhani, Shagun and Zhang, Amy and Pineau, Joelle},
  booktitle={International Conference on Machine Learning (ICML)},
  pages={9767--9779},
  year={2021},
  organization={PMLR}
}

@inproceedings{Ref_Wang_Dueling_ICML16,
  title={Dueling network architectures for deep reinforcement learning},
  author={Wang, Ziyu and Schaul, Tom and Hessel, Matteo and Hasselt, Hado and Lanctot, Marc and Freitas, Nando},
  booktitle={International Conference on Machine Learning (ICML)},
  pages={1995--2003},
  year={2016},
  organization={PMLR}
}

@ARTICLE{MNIST,
  author={Deng, Li},
  journal={IEEE Signal Processing Magazine}, 
  title={The MNIST Database of Handwritten Digit Images for Machine Learning Research [Best of the Web]}, 
  year={2012},
  volume={29},
  number={6},
  pages={141-142},
  doi={10.1109/MSP.2012.2211477}}

@article{FMNIST,
  title={Fashion-mnist: a novel image dataset for benchmarking machine learning algorithms},
  author={Xiao, Han and Rasul, Kashif and Vollgraf, Roland},
  journal={arXiv preprint arXiv:1708.07747},
  year={2017}
}

@article{krizhevsky2014cifar,
  title={The CIFAR-10 dataset},
  author={Krizhevsky, Alex and Nair, Vinod and Hinton, Geoffrey and others},
  journal={online: http://www.cs. toronto.edu/kriz/cifar.html},
  volume={55},
  number={5},
  pages={2},
  year={2014}
}

@article{parisotto2016actormimicdeepmultitasktransfer,
  title={Actor-mimic: Deep multitask and transfer reinforcement learning},
  author={Parisotto, Emilio and Ba, Jimmy Lei and Salakhutdinov, Ruslan},
  journal={arXiv preprint arXiv:1511.06342},
  year={2015}
}

@article{rusu2016policydistillation,
  title={Policy distillation},
  author={Rusu, Andrei A and Colmenarejo, Sergio Gomez and Gulcehre, Caglar and Desjardins, Guillaume and Kirkpatrick, James and Pascanu, Razvan and Mnih, Volodymyr and Kavukcuoglu, Koray and Hadsell, Raia},
  journal={arXiv preprint arXiv:1511.06295},
  year={2015}
}

@ArtifactSoftware{R,
    title = {R: A Language and Environment for Statistical Computing},
    author = {{R Core Team}},
    organization = {R Foundation for Statistical Computing},
    address = {Vienna, Austria},
    year = {2019},
    url = {https://www.R-project.org/},
}
